\documentclass[10pt,twocolumn,letterpaper]{article}

\usepackage{cvpr}              
\usepackage[accsupp]{axessibility}

\usepackage{graphicx}
\usepackage{makecell}
\usepackage{booktabs}
\usepackage{adjustbox}
\usepackage{float}
\definecolor{gray60}{gray}{0.6}
\definecolor{baselineblue}{RGB}{230,245,255}
\usepackage{xcolor}
\usepackage{tcolorbox}

\definecolor{cvprblue}{rgb}{0.21,0.49,0.74}
\usepackage[pagebackref,breaklinks,colorlinks,allcolors=cvprblue]{hyperref}

\def\paperID{***} 
\def\confName{CVPR}
\def\confYear{2026}

\title{A Comprehensive Study on Visual Token Redundancy for Discrete Diffusion-based Multimodal Large Language Models}

\author{
  Duo Li$^{1}$\thanks{Equal Contribution} \quad
  Zuhao Yang$^{1}$\footnotemark[1] \quad
  Xiaoqin Zhang$^{2}$ \quad
  Ling Shao$^{3}$ \quad
  Shijian Lu$^{1}$\thanks{Corresponding Author} \\
  \vspace{0.8em}
  $^{1}$CCDS, NTU, Singapore \quad
  $^{2}$CCST, ZJUT, China \quad
  $^{3}$Terminus AI Lab, UCAS, China \\
  \vspace{0.5em}
  Email Contact: \texttt{\{duo001, yang0756\}@e.ntu.edu.sg}
}

\begin{document}

\maketitle
\begin{figure*}[t]
\begin{center}
  \includegraphics[width=\textwidth,trim=0 10 0 6,clip]{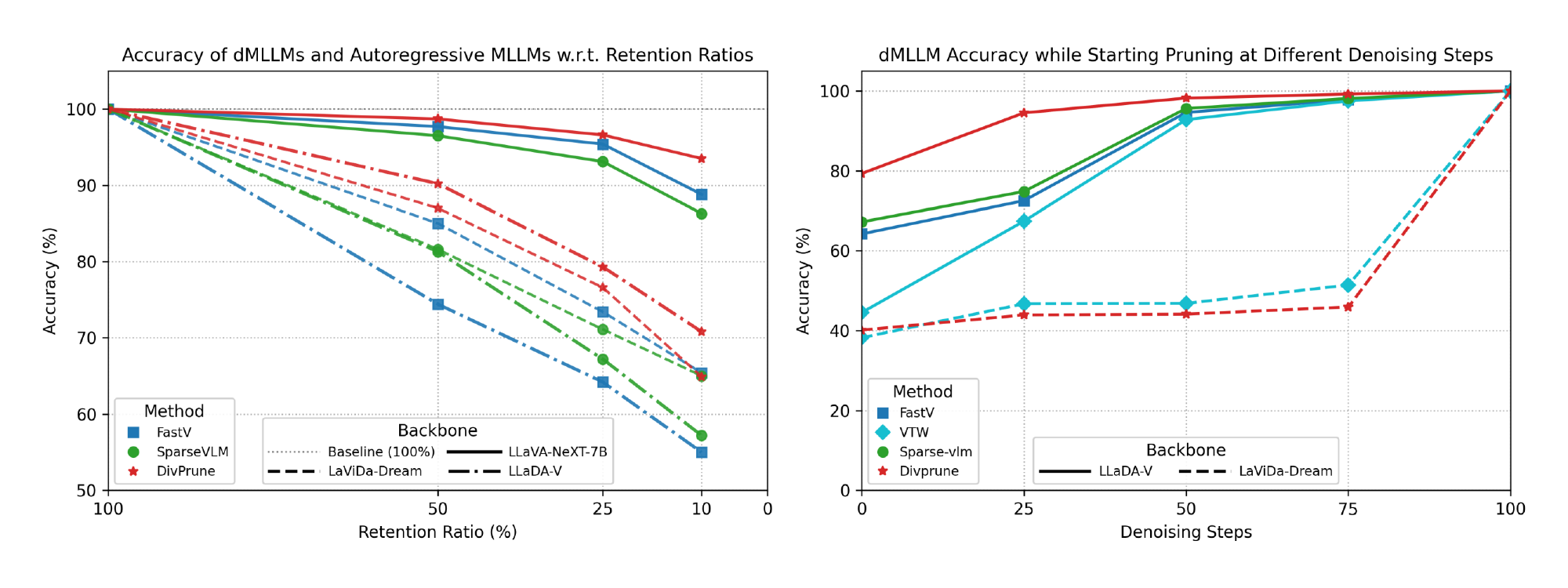}
  \captionof{figure}{\textbf{\textsc{Left:} Visual token pruning affects dMLLMs much more than autoregressive MLLMs.} Three representative pruning methods \cite{chen2024fastv, zhang2024sparsevlm, alvar2025divprune} are applied to LLaVA-NeXT \cite{liu2024llavanext} (solid lines), LaViDa-Dream \cite{li2025lavida}, and LLaDA-V \cite{you2025lladav} (dotted lines). \textbf{\textsc{Right:} Performance vs.\ pruning start step.} Unlike AR-to-diffusion LaViDa-Dream, the from-scratch LLaDA-V consistently recovers pruning-induced information loss, achieving much higher accuracy when pruning starts at different denoising steps (projected to 0--100 for visualization).
  }
  \label{fig:retention-accuracy}
\end{center}
\end{figure*}

\begin{abstract}
Discrete diffusion-based multimodal large language models (dMLLMs) have emerged as a promising alternative to autoregressive MLLMs thanks to their advantages in parallel decoding and bidirectional context modeling, but most existing dMLLMs incur significant computational overhead during inference due to the full-sequence attention computation in each denoising step. Pioneering studies attempt to resolve this issue from a modality-agnostic perspective via key–value cache optimization or efficient sampling but most of them overlook modality-specific visual token redundancy. 
In this work, we conduct a comprehensive study on how visual token redundancy evolves with different dMLLM architectures and tasks and how visual token pruning affects dMLLM responses and efficiency.
Specifically, our study reveals that visual redundancy emerges only in from-scratch dMLLMs while handling long-answer tasks.
In addition, we validate that visual token pruning introduces non-negligible information loss in dMLLMs and only from-scratch dMLLMs can recover the lost information progressively during late denoising steps.
Furthermore, our study shows that layer-skipping is promising for accelerating AR-to-diffusion dMLLMs, whereas progressive or late-step pruning is more effective for from-scratch dMLLMs.
Overall, this work offers a new perspective on efficiency optimization for dMLLMs, greatly advancing their applicability across various multimodal understanding tasks. The code is available at: \url{https://github.com/Yrdal3910/dMLLM-Visual-Token-Redundancy-Analysis}.
\end{abstract}    
\section{Introduction}
\label{sec:intro}

Discrete diffusion-based large language models (dLLMs) have recently emerged and advanced rapidly thanks to their promising potential in both efficiency and performance~\cite{khanna2025mercury}.
Most existing dLLMs are trained with two representative approaches: 1) \emph{from-scratch diffusion training} \cite{ye2023diffusionllms, nie2025llada, zhu2025llada15} that trains standard Transformers \cite{vaswani2017transformers} from scratch to directly learn the denoising-based text generation process and 2) \emph{AR-to-diffusion adaptation} \cite{gong2024DiffuGPT, ye2025dream, tae2025tess} that converts pretrained autoregressive large language models (LLMs) into diffusion-based generators by further training them with masking-based diffusion objectives.
Leveraging the advances in dLLMs, several discrete diffusion-based multimodal LLMs (dMLLMs) \cite{you2025lladav, li2025lavida, yu2025dimple} have been developed and achieved competitive performance as compared with autoregressive multimodal large language models (MLLMs) of similar scales.

However, most open-source dMLLMs involve intensive computation during inference, leading to a much longer inference process as compared with autoregressive MLLMs \cite{ma2025dkvcache}. Several studies attempt to address this issue via different optimization strategies such as key–value cache optimization \cite{liu2025dllmcache,wu2025fastdllm,song2025sparsedllm}, efficient token sampling \cite{wu2025fastdllm,li2025prophet, wei2025slowfast,huang2025pcsampler}, and variable-length generation \cite{wu2025dreamon, li2025daedel, kim2025flexmdm}. However, most of these studies tackle the problem from a modality-agnostic perspective without considering the redundancy of visual tokens. Given that visual token redundancy and visual token pruning have been widely studied for autoregressive MLLMs \cite{liang2022eviT, ryoo2021tokenlearner, shang2024llavaprumerge}, two straightforward questions naturally come to our mind: 1) whether visual token redundancy exists in prevalent dMLLMs, and if so, 2) how visual token pruning affects the inference speed and accuracy of prevalent dMLLMs. A direct extension to both questions is how those pruning approaches that have achieved great success for autoregressive MLLMs perform in dMLLMs.

We perform a systematic study on whether visual token redundancy exists, how it affects dMLLMs, and how it could guide visual token pruning in dMLLMs. The study focuses on two prevalent dMLLM backbones, namely, from-scratch dMLLMs (e.g., LLaDA-V \cite{you2025lladav}) and AR-to-diffusion dMLLMs (e.g., LaViDa-Dream \cite{li2025lavida}), as well as six representative token pruning techniques that achieve impressive performance in autoregressive MLLMs. We first apply visual token pruning to the initial denoising steps, and the substantial performance drops in both dMLLM backbones (see \Cref{fig:retention-accuracy} left) indicate the essential role of visual tokens at the initial denoising stage. On the other hand, the two backbones exhibit very divergent behaviors while applying visual token pruning to the following denoising steps (see \Cref{fig:retention-accuracy} right). Specifically, AR-to-diffusion dMLLMs still suffer from severe performance drops for both short-answer and long-answer tasks, while from-scratch dMLLMs suffer from a similar performance drop for short-answer tasks but achieve only minimal performance drops 
for long-answer tasks. In addition, we examine why the two dMLLM backbones perform very differently for long-answer tasks.
Our study shows that visual token pruning leads to much more information loss in dMLLMs than in MLLMs, while from-scratch dMLLMs can better restore the pruning-induced information loss via the subsequent iterative bidirectional denoising and refinement, explaining their minimal performance drops. 
Further, our study shows that both attention scores during inference and step-wise output logits can be effective indicators to guide visual token pruning in dMLLMs. \looseness=-1

The contributions of this work are three-fold.
First, we conduct the first comprehensive study of visual token redundancy in dMLLMs, revealing that AR-to-diffusion dMLLMs suffer consistent and clear performance drops under pruning while from-scratch dMLLMs remain resilient on long-answer tasks.
Second, we show that pruning causes much greater information loss in dMLLMs than in MLLMs, and that from-scratch dMLLMs can mitigate such loss via iterative bidirectional denoising whereas AR-to-diffusion dMLLMs lack this restoration ability.
Third, we offer practical guidance: layer-skipping suits AR-to-diffusion dMLLMs, while progressive or late-step pruning works better for from-scratch architectures, with both attention scores and logit signals serving as reliable pruning indicators.

\section{Related Work}
\label{sec:related_work}

\subsection{Visual Token Compression for MLLMs}
\label{sec:related_work_vtc}

MLLMs typically encode visual inputs into a large set of visual tokens.
These visual tokens are inherently sparse in representing effective information compared to textual tokens \cite{yang2025explainingvisualredundancy, shang2024llavaprumerge}, thereby incurring considerable yet unnecessary computational cost.
To combat this inefficiency, existing visual token compression methods attempt to prune redundant visual tokens or skip certain computational stages. 
These approaches can be broadly categorized into four groups:
1) transformation-based compression, which weakens the spatial and temporal redundancy inherent in visual inputs by applying pixel unshuffling \cite{wang2024qwen2, bai2025qwen25vl, gao2024miniinternvl, chen2024internvl25}, pooling \cite{zhang2024llavavideo, yao2024deco}, and convolutional transformations \cite{cheng2024videollama2, chu2023mobilevlm};
2) similarity-based compression, which prunes or merges redundant tokens according to inter-token similarity or distance metrics, typically measured via cosine similarity \cite{bolya2022tome, alvar2025divprune, yang2025topv}; 
3) attention-based compression, which employs various attention scores as the importance metric to guide token pruning in either the vision encoder \cite{liu2025globalcom, yang2025visionzip} or the LLM decoder \cite{chen2024fastv, xing2024pyramiddrop};
and 4) query-based compression, which leverages query prompts to guide the visual token compression and can be further classified into token distillation \cite{bai2023qwen, zhu2023minigpt4} and cross-modal selection strategies \cite{zhang2024sparsevlm, ruiz2024trim}.

\subsection{Acceleration Techniques for dMLLMs} 
\label{sec:related_work_acce_dmllms}

Unlike LLMs that generate tokens sequentially with cached context, dLLMs perform iterative denoising across numerous steps, each involving bidirectional full-sequence attention and softmax computations, which collectively lead to significant slowdown in inference.
To overcome these bottlenecks, recent approaches have accelerated dLLMs from three perspectives: 
1) Optimizing key-value (KV) cache utilization\textemdash{}reusing KV caches across denoising steps \cite{wu2025fastdllm, liu2025dllmcache} and reducing cached representations within each step \cite{song2025sparsedllm, chen2025dpad};
2) Designing efficient sampling algorithms\textemdash{}accelerating inference via dynamic and parallel decoding \cite{wu2025fastdllm, li2025prophet, wei2025slowfast, huang2025pcsampler};
and 3) Adopting variable-length generation\textemdash{}enabling the model to dynamically adjust the output sequence length during denoising \cite{wu2025dreamon, li2025daedel, kim2025flexmdm}.
While these methods can be naturally inherited by dMLLMs, a modality-specific yet overlooked factor is visual token redundancy.
Our work aims to provide the first systematic study of this underexplored aspect and discuss its potential optimization directions.

\section{From Autoregression to Diffusion: Modeling and Complexity}
\label{sec:preliminary}

\subsection{Model Foundation}
\label{sec:preliminary_formulation}

\paragraph{Foundation of MLLMs.}
MLLMs adopt autoregressive modeling to define the model distribution over the output sequence \cite{wu2023mllmsurvey}.
Given a multimodal context $x$ (e.g., image–text pair) and target sequence $y = (y^1, \dots, y^L)$, the model factorizes the conditional likelihood via the chain rule of probability:
\begin{equation}
p_{\theta}(y|x) = \prod_{t=1}^{L} p_{\theta}(y^{t} \mid x, y^{<t}),
\label{eq:ar_model}
\end{equation}
where $y^{<t} = (y^1, \dots, y^{t-1})$, and $L$ is the total length of the target sequence.
The parameters $\theta$ are optimized by maximum likelihood estimation:
\begin{equation}
\mathcal{L}_{\text{AR}}(\theta)
= - \mathbb{E}_{(x,y) \sim p_{\text{data}}}
\sum_{t=1}^{L} \log p_{\theta}(y^{t} \mid x, y^{<t}).
\end{equation}
This formulation enforces a \emph{causal and unidirectional} dependency, 
where each token prediction conditions on all preceding tokens and the multimodal input context.

\paragraph{Foundation of dMLLMs.}
In contrast, dMLLMs parameterize a denoising Markov chain over discrete latent sequences \cite{nie2025llada, ye2025dream}.
Let $y_0$ denote the clean output sequence and $y_t$ represent its masked version at diffusion step $t \in [1,T]$.
The forward noising process $q_{t|0}(y_t \mid y_0)$ gradually injects noise or masks into $y_0$, producing a sequence of intermediate noisy states $\{y_t\}_{t=1}^{T}$.
The reverse denoising process is modeled by a parameterized consistency kernel:
\begin{equation}
p_{\theta}(y_{t-1} \mid y_t, x),
\end{equation}
which approximates the ideal reverse kernel that satisfies Chapman–Kolmogorov consistency \cite{Kolmogorov1931ChapmanKolmogorovequation}.
The overall model distribution is given by marginalizing over all intermediate states:
\begin{equation}
p_{\theta}(y_0 \mid x)
= \sum_{y_{1:T}} p(y_T)\!\!\prod_{t=1}^{T} p_{\theta}(y_{t-1} \mid y_t, x),
\end{equation}
where $p(y_T)$ is the prior distribution (e.g., a fully masked state where $p(y_T)=\delta(y_T=M)$).
The parameters $\theta$ are optimized by minimizing a negative log-likelihood upper bound:
\begin{equation}
\mathcal{L}_{\text{Diff}}(\theta)
= -\, \mathbb{E}_{y_0, t, y_t} \Big[
w(t)\!\!\sum_{i=1}^{L} \mathbf{1}[y_t^i = M]
\log p_{\theta}(y_0^i \mid y_t, x)
\Big],
\label{eq:df_loss}
\end{equation}
where $M$ denotes the masked token, and $w(t)$ is a weighting function derived from the diffusion noise schedule.
Unlike causal factorization (Eq.~\eqref{eq:ar_model}), Eq.~\eqref{eq:df_loss} defines a non-causal and iterative denoising process that jointly models all tokens in each step, enabling \emph{bidirectional} contextual reasoning across modalities.

\begin{table*}[!t]
\centering

\resizebox{\linewidth}{!}{%
\begin{tabular}{l|*{8}{r@{}l}|r@{}l}
\toprule[1pt]
\multicolumn{1}{c|}{\textbf{Method}} & \multicolumn{2}{c}{\textbf{MME} \cite{Fu2023mme}} & \multicolumn{2}{c}{\textbf{SQA} \cite{lu2022sqa}} & \multicolumn{2}{c}{\textbf{GQA} \cite{hudson2019gqa}} & \multicolumn{2}{c}{\textbf{POPE} \cite{li2023pope}} & 
\multicolumn{2}{c}{\textbf{MMB} \cite{liu2024mmbench}} & \multicolumn{2}{c}{\textbf{TVQA} \cite{singh2019textvqa}} & \multicolumn{2}{c}{\textbf{CQA} \cite{masry2022chartqa}} & \multicolumn{2}{c}{\textbf{MMMUP} \cite{yue2024mmmupro}} & \multicolumn{2}{c}{\textbf{Avg.}}\\
\midrule

LLaDA-V \cite{you2025lladav}
& 1994.4 & \makecell[l]{\mbox{}\\ \textcolor{gray}{5:47}}
& 89.8   & \makecell[l]{\mbox{}\\ \textcolor{gray}{3:07}}
& 52.1   & \makecell[l]{\mbox{}\\ \textcolor{gray}{22:25}}
& 87.3   & \makecell[l]{\mbox{}\\ \textcolor{gray}{15:45}}
& 83.3   & \makecell[l]{\mbox{}\\ \textcolor{gray}{7:24}}
& 31.7   & \makecell[l]{\mbox{}\\ \textcolor{gray}{10:49}}
& 77.7   & \makecell[l]{\mbox{}\\ \textcolor{gray}{10:08}}
& 18.7   & \makecell[l]{\mbox{}\\ \textcolor{gray}{4:19}}
& 100.0\%& \makecell[l]{\mbox{}\\ \textcolor{gray}{79:44}}
\\

\midrule
\multicolumn{19}{c}{\textbf{Retention Ratio = 50\%}}\\
\midrule
ToMe (ICLR'23) \cite{bolya2022tome}
& 1918.7 & \raisebox{-0.35ex}{\scriptsize\textcolor{ForestGreen}{$\uparrow\,4.9\%$}} 
& 86.9   & \raisebox{-0.35ex}{\scriptsize\textcolor{ForestGreen}{$\uparrow\,3.7\%$}}
& 50.0   & \raisebox{-0.35ex}{\scriptsize\textcolor{ForestGreen}{$\uparrow\,4.7\%$}}
& 87.4   & \raisebox{-0.35ex}{\scriptsize\textcolor{ForestGreen}{$\uparrow\,6.1\%$}}
& 79.3   & \raisebox{-0.35ex}{\scriptsize\textcolor{ForestGreen}{$\uparrow\,13.7\%$}}
& 29.9   & \raisebox{-0.35ex}{\scriptsize\textcolor{BrickRed}{$\downarrow\,0.9\%$}}
& 53.2   & \raisebox{-0.35ex}{\scriptsize\textcolor{BrickRed}{$\downarrow\,32.1\%$}}
& 15.7   & \raisebox{-0.35ex}{\scriptsize\textcolor{BrickRed}{$\downarrow\,8.9\%$}}
& 91.3\% & \raisebox{-0.35ex}{\scriptsize\textcolor{BrickRed}{$\downarrow\,0.4\%$}}
\\
FastV (ECCV'24) \cite{chen2024fastv}
& 1655.8 & \raisebox{-0.35ex}{\scriptsize\textcolor{ForestGreen}{$\uparrow\,15.3\%$}}
& 78.3   & \raisebox{-0.35ex}{\scriptsize\textcolor{BrickRed}{$\downarrow\,3.2\%$}}
& 43.5   & \raisebox{-0.35ex}{\scriptsize\textcolor{BrickRed}{$\downarrow\,0.7\%$}}
& 82.1   & \raisebox{-0.35ex}{\scriptsize\textcolor{BrickRed}{$\downarrow\,6.6\%$}}
& 67.6   & \raisebox{-0.35ex}{\scriptsize\textcolor{ForestGreen}{$\uparrow\,7.2\%$}}
& 25.6   & \raisebox{-0.35ex}{\scriptsize\textcolor{BrickRed}{$\downarrow\,1.4\%$}}
& 15.6   & \raisebox{-0.35ex}{\scriptsize\textcolor{BrickRed}{$\downarrow\,26.5\%$}}
& 12.1   & \raisebox{-0.35ex}{\scriptsize\textcolor{ForestGreen}{$\uparrow\,0.8\%$}}
& 74.4\% & \raisebox{-0.35ex}{\scriptsize\textcolor{BrickRed}{$\downarrow\,3.4\%$}}
\\
TRIM (COLING'25) \cite{ruiz2024trim}
& 1590.4 & \raisebox{-0.35ex}{\scriptsize\textcolor{ForestGreen}{$\uparrow\,40.3\%$}}
& 82.2   & \raisebox{-0.35ex}{\scriptsize\textcolor{ForestGreen}{$\uparrow\,22.5\%$}}
& 42.8   & \raisebox{-0.35ex}{\scriptsize\textcolor{ForestGreen}{$\uparrow\,20.4\%$}}
& 77.3   & \raisebox{-0.35ex}{\scriptsize\textcolor{ForestGreen}{$\uparrow\,35.9\%$}}
& 75.9   & \raisebox{-0.35ex}{\scriptsize\textcolor{ForestGreen}{$\uparrow\,34.9\%$}}
& 20.9   & \raisebox{-0.35ex}{\scriptsize\textcolor{ForestGreen}{$\uparrow\,42.1\%$}}
& 32.7   & \raisebox{-0.35ex}{\scriptsize\textcolor{ForestGreen}{$\uparrow\,47.9\%$}}
& 12.0   & \raisebox{-0.35ex}{\scriptsize\textcolor{ForestGreen}{$\uparrow\,25.9\%$}}
& 75.6\% & \raisebox{-0.35ex}{\scriptsize\textcolor{ForestGreen}{$\uparrow\,33.1\%$}}
\\
SparseVLM (ICML'25) \cite{zhang2024sparsevlm}
& 1730.1 & \raisebox{-0.35ex}{\scriptsize\textcolor{BrickRed}{$\downarrow\,9.5\%$}}
& 80.6   & \raisebox{-0.35ex}{\scriptsize\textcolor{BrickRed}{$\downarrow\,15.5\%$}}
& 47.8   & \raisebox{-0.35ex}{\scriptsize\textcolor{BrickRed}{$\downarrow\,20.5\%$}}
& 83.1   & \raisebox{-0.35ex}{\scriptsize\textcolor{BrickRed}{$\downarrow\,22.0\%$}}
& 73.2   & \raisebox{-0.35ex}{\scriptsize\textcolor{BrickRed}{$\downarrow\,5.9\%$}}
& 23.1   & \raisebox{-0.35ex}{\scriptsize\textcolor{BrickRed}{$\downarrow\,39.4\%$}}
& 38.0   & \raisebox{-0.35ex}{\scriptsize\textcolor{BrickRed}{$\downarrow\,72.0\%$}}
& 14.5   & \raisebox{-0.35ex}{\scriptsize\textcolor{BrickRed}{$\downarrow\,42.9\%$}}
& 81.3\% & \raisebox{-0.35ex}{\scriptsize\textcolor{BrickRed}{$\downarrow\,28.8\%$}}
\\
DivPrune (CVPR'25) \cite{alvar2025divprune}
& 1793.7 & \raisebox{-0.35ex}{\scriptsize\textcolor{ForestGreen}{$\uparrow\,12.1\%$}}
& 82.3   & \raisebox{-0.35ex}{\scriptsize\textcolor{BrickRed}{$\downarrow\,10.7\%$}}
& 48.3   & \raisebox{-0.35ex}{\scriptsize\textcolor{BrickRed}{$\downarrow\,25.7\%$}}
& 86.1   & \raisebox{-0.35ex}{\scriptsize\textcolor{BrickRed}{$\downarrow\,6.3\%$}}
& 74.6   & \raisebox{-0.35ex}{\scriptsize\textcolor{BrickRed}{$\downarrow\,1.4\%$}}
& 29.5   & \raisebox{-0.35ex}{\scriptsize\textcolor{ForestGreen}{$\uparrow\,5.1\%$}}
& 64.8   & \raisebox{-0.35ex}{\scriptsize\textcolor{ForestGreen}{$\uparrow\,18.4\%$}}
& 15.5   & \raisebox{-0.35ex}{\scriptsize\textcolor{BrickRed}{$\downarrow\,1.5\%$}}
& 90.2\% & \raisebox{-0.35ex}{\scriptsize\textcolor{BrickRed}{$\downarrow\,5.2\%$}}
\\
\midrule
\multicolumn{19}{c}{\textbf{Retention Ratio = 25\%}}\\
\midrule
ToMe (ICLR'23) \cite{bolya2022tome}
& 1799.6 & \raisebox{-0.35ex}{\scriptsize\textcolor{ForestGreen}{$\uparrow\,20.2\%$}}
& 84.6   & \raisebox{-0.35ex}{\scriptsize\textcolor{ForestGreen}{$\uparrow\,26.2\%$}}
& 48.4   & \raisebox{-0.35ex}{\scriptsize\textcolor{ForestGreen}{$\uparrow\,28.0\%$}}
& 87.1   & \raisebox{-0.35ex}{\scriptsize\textcolor{ForestGreen}{$\uparrow\,24.0\%$}}
& 78.3   & \raisebox{-0.35ex}{\scriptsize\textcolor{ForestGreen}{$\uparrow\,27.5\%$}}
& 28.8   & \raisebox{-0.35ex}{\scriptsize\textcolor{ForestGreen}{$\uparrow\,21.3\%$}}
& 45.6   & \raisebox{-0.35ex}{\scriptsize\textcolor{ForestGreen}{$\uparrow\,7.2\%$}}
& 14.3   & \raisebox{-0.35ex}{\scriptsize\textcolor{BrickRed}{$\downarrow\,0.8\%$}}
& 87.1\% & \raisebox{-0.35ex}{\scriptsize\textcolor{ForestGreen}{$\uparrow\,21.4\%$}}
\\
FastV (ECCV'24) \cite{chen2024fastv}
& 1374.1 & \raisebox{-0.35ex}{\scriptsize\textcolor{ForestGreen}{$\uparrow\,34.9\%$}}
& 74.2   & \raisebox{-0.35ex}{\scriptsize\textcolor{ForestGreen}{$\uparrow\,15.0\%$}}
& 39.1   & \raisebox{-0.35ex}{\scriptsize\textcolor{ForestGreen}{$\uparrow\,19.3\%$}}
& 74.6   & \raisebox{-0.35ex}{\scriptsize\textcolor{ForestGreen}{$\uparrow\,17.9\%$}}
& 58.2   & \raisebox{-0.35ex}{\scriptsize\textcolor{ForestGreen}{$\uparrow\,21.2\%$}}
& 18.0   & \raisebox{-0.35ex}{\scriptsize\textcolor{ForestGreen}{$\uparrow\,25.9\%$}}
& 10.8   & \raisebox{-0.35ex}{\scriptsize\textcolor{ForestGreen}{$\uparrow\,31.6\%$}}
& 11.5   & \raisebox{-0.35ex}{\scriptsize\textcolor{ForestGreen}{$\uparrow\,23.2\%$}}
& 64.2\% & \raisebox{-0.35ex}{\scriptsize\textcolor{ForestGreen}{$\uparrow\,22.8\%$}}
\\
TRIM (COLING'25) \cite{ruiz2024trim}
& 1243.5 & \raisebox{-0.35ex}{\scriptsize\textcolor{ForestGreen}{$\uparrow\,47.0\%$}}
& 78.2   & \raisebox{-0.35ex}{\scriptsize\textcolor{ForestGreen}{$\uparrow\,25.1\%$}}
& 38.0   & \raisebox{-0.35ex}{\scriptsize\textcolor{ForestGreen}{$\uparrow\,31.7\%$}}
& 68.0   & \raisebox{-0.35ex}{\scriptsize\textcolor{ForestGreen}{$\uparrow\,39.9\%$}}
& 68.7   & \raisebox{-0.35ex}{\scriptsize\textcolor{ForestGreen}{$\uparrow\,39.4\%$}}
& 16.2   & \raisebox{-0.35ex}{\scriptsize\textcolor{ForestGreen}{$\uparrow\,45.8\%$}}
& 26.2   & \raisebox{-0.35ex}{\scriptsize\textcolor{ForestGreen}{$\uparrow\,53.3\%$}}
& 12.3   & \raisebox{-0.35ex}{\scriptsize\textcolor{ForestGreen}{$\uparrow\,29.3\%$}}
& 66.7\% & \raisebox{-0.35ex}{\scriptsize\textcolor{ForestGreen}{$\uparrow\,39.4\%$}}
\\
SparseVLM (ICML'25) \cite{zhang2024sparsevlm}
& 1437.1 & \raisebox{-0.35ex}{\scriptsize\textcolor{ForestGreen}{$\uparrow\,13.3\%$}}
& 75.1   & \raisebox{-0.35ex}{\scriptsize\textcolor{ForestGreen}{$\uparrow\,19.3\%$}}
& 43.2   & \raisebox{-0.35ex}{\scriptsize\textcolor{ForestGreen}{$\uparrow\,4.6\%$}}
& 76.8   & \raisebox{-0.35ex}{\scriptsize\textcolor{ForestGreen}{$\uparrow\,2.2\%$}}
& 62.1   & \raisebox{-0.35ex}{\scriptsize\textcolor{ForestGreen}{$\uparrow\,18.7\%$}}
& 14.7   & \raisebox{-0.35ex}{\scriptsize\textcolor{BrickRed}{$\downarrow\,2.3\%$}}
& 16.4   & \raisebox{-0.35ex}{\scriptsize\textcolor{ForestGreen}{$\uparrow\,3.5\%$}}
& 12.9   & \raisebox{-0.35ex}{\scriptsize\textcolor{BrickRed}{$\downarrow\,13.5\%$}}
& 67.2\% & \raisebox{-0.35ex}{\scriptsize\textcolor{ForestGreen}{$\uparrow\,4.6\%$}}
\\
DivPrune (CVPR'25) \cite{alvar2025divprune}
& 1574.4 & \raisebox{-0.35ex}{\scriptsize\textcolor{ForestGreen}{$\uparrow\,25.4\%$}}
& 80.5   & \raisebox{-0.35ex}{\scriptsize\textcolor{ForestGreen}{$\uparrow\,5.3\%$}}
& 43.9   & \raisebox{-0.35ex}{\scriptsize\textcolor{ForestGreen}{$\uparrow\,10.0\%$}}
& 84.2   & \raisebox{-0.35ex}{\scriptsize\textcolor{ForestGreen}{$\uparrow\,7.7\%$}}
& 70.2   & \raisebox{-0.35ex}{\scriptsize\textcolor{ForestGreen}{$\uparrow\,7.9\%$}}
& 26.5   & \raisebox{-0.35ex}{\scriptsize\textcolor{ForestGreen}{$\uparrow\,20.0\%$}}
& 33.7   & \raisebox{-0.35ex}{\scriptsize\textcolor{ForestGreen}{$\uparrow\,40.3\%$}}
& 13.8   & \raisebox{-0.35ex}{\scriptsize\textcolor{ForestGreen}{$\uparrow\,15.8\%$}}
& 79.3\% & \raisebox{-0.35ex}{\scriptsize\textcolor{ForestGreen}{$\uparrow\,15.8\%$}}
\\
\midrule
\multicolumn{19}{c}{\textbf{Retention Ratio = 10\%}}\\
\midrule
ToMe (ICLR'23) \cite{bolya2022tome}
& 1664.0 & \raisebox{-0.35ex}{\scriptsize\textcolor{ForestGreen}{$\uparrow\,41.8\%$}}
& 82.7   & \raisebox{-0.35ex}{\scriptsize\textcolor{ForestGreen}{$\uparrow\,30.5\%$}}
& 45.5   & \raisebox{-0.35ex}{\scriptsize\textcolor{ForestGreen}{$\uparrow\,36.7\%$}}
& 85.8   & \raisebox{-0.35ex}{\scriptsize\textcolor{ForestGreen}{$\uparrow\,35.8\%$}}
& 75.2   & \raisebox{-0.35ex}{\scriptsize\textcolor{ForestGreen}{$\uparrow\,43.5\%$}}
& 26.4   & \raisebox{-0.35ex}{\scriptsize\textcolor{ForestGreen}{$\uparrow\,37.4\%$}}
& 33.0   & \raisebox{-0.35ex}{\scriptsize\textcolor{ForestGreen}{$\uparrow\,38.8\%$}}
& 14.0   & \raisebox{-0.35ex}{\scriptsize\textcolor{ForestGreen}{$\uparrow\,5.4\%$}}
& 81.5\% & \raisebox{-0.35ex}{\scriptsize\textcolor{ForestGreen}{$\uparrow\,36.0\%$}}
\\
FastV (ECCV'24) \cite{chen2024fastv}
& 1192.7 & \raisebox{-0.35ex}{\scriptsize\textcolor{ForestGreen}{$\uparrow\,40.9\%$}}
& 73.4   & \raisebox{-0.35ex}{\scriptsize\textcolor{ForestGreen}{$\uparrow\,22.5\%$}}
& 35.7   & \raisebox{-0.35ex}{\scriptsize\textcolor{ForestGreen}{$\uparrow\,21.3\%$}}
& 63.7   & \raisebox{-0.35ex}{\scriptsize\textcolor{ForestGreen}{$\uparrow\,28.7\%$}}
& 46.0   & \raisebox{-0.35ex}{\scriptsize\textcolor{ForestGreen}{$\uparrow\,27.0\%$}}
& 9.1    & \raisebox{-0.35ex}{\scriptsize\textcolor{ForestGreen}{$\uparrow\,31.1\%$}}
& 9.5    & \raisebox{-0.35ex}{\scriptsize\textcolor{ForestGreen}{$\uparrow\,48.5\%$}}
& 11.3   & \raisebox{-0.35ex}{\scriptsize\textcolor{ForestGreen}{$\uparrow\,25.9\%$}}
& 55.0\% & \raisebox{-0.35ex}{\scriptsize\textcolor{ForestGreen}{$\uparrow\,29.8\%$}}
\\
TRIM (COLING'25) \cite{ruiz2024trim}
& 1092.3 & \raisebox{-0.35ex}{\scriptsize\textcolor{ForestGreen}{$\uparrow\,47.6\%$}}
& 75.6   & \raisebox{-0.35ex}{\scriptsize\textcolor{ForestGreen}{$\uparrow\,29.4\%$}}
& 35.3   & \raisebox{-0.35ex}{\scriptsize\textcolor{ForestGreen}{$\uparrow\,39.4\%$}}
& 60.7   & \raisebox{-0.35ex}{\scriptsize\textcolor{ForestGreen}{$\uparrow\,49.0\%$}}
& 59.7   & \raisebox{-0.35ex}{\scriptsize\textcolor{ForestGreen}{$\uparrow\,45.7\%$}}
& 11.1   & \raisebox{-0.35ex}{\scriptsize\textcolor{ForestGreen}{$\uparrow\,48.2\%$}}
& 13.3   & \raisebox{-0.35ex}{\scriptsize\textcolor{ForestGreen}{$\uparrow\,56.4\%$}}
& 11.4   & \raisebox{-0.35ex}{\scriptsize\textcolor{ForestGreen}{$\uparrow\,32.4\%$}}
& 57.6\% & \raisebox{-0.35ex}{\scriptsize\textcolor{ForestGreen}{$\uparrow\,45.1\%$}}
\\
SparseVLM (ICML'25) \cite{zhang2024sparsevlm}
& 1267.1 & \raisebox{-0.35ex}{\scriptsize\textcolor{ForestGreen}{$\uparrow\,23.3\%$}}
& 73.4   & \raisebox{-0.35ex}{\scriptsize\textcolor{ForestGreen}{$\uparrow\,26.2\%$}}
& 37.3   & \raisebox{-0.35ex}{\scriptsize\textcolor{ForestGreen}{$\uparrow\,18.8\%$}}
& 67.6   & \raisebox{-0.35ex}{\scriptsize\textcolor{ForestGreen}{$\uparrow\,18.9\%$}}
& 46.9   & \raisebox{-0.35ex}{\scriptsize\textcolor{ForestGreen}{$\uparrow\,21.4\%$}}
& 9.2    & \raisebox{-0.35ex}{\scriptsize\textcolor{ForestGreen}{$\uparrow\,11.1\%$}}
& 10.6   & \raisebox{-0.35ex}{\scriptsize\textcolor{ForestGreen}{$\uparrow\,27.5\%$}}
& 12.0   & \raisebox{-0.35ex}{\scriptsize\textcolor{BrickRed}{$\downarrow\,1.9\%$}}
& 57.2\% & \raisebox{-0.35ex}{\scriptsize\textcolor{ForestGreen}{$\uparrow\,18.6\%$}}
\\
DivPrune (CVPR'25) \cite{alvar2025divprune}
& 1468.4 & \raisebox{-0.35ex}{\scriptsize\textcolor{ForestGreen}{$\uparrow\,33.7\%$}}
& 78.4   & \raisebox{-0.35ex}{\scriptsize\textcolor{ForestGreen}{$\uparrow\,8.0\%$}}
& 40.3   & \raisebox{-0.35ex}{\scriptsize\textcolor{ForestGreen}{$\uparrow\,48.9\%$}}
& 80.7   & \raisebox{-0.35ex}{\scriptsize\textcolor{ForestGreen}{$\uparrow\,11.5\%$}}
& 62.8   & \raisebox{-0.35ex}{\scriptsize\textcolor{ForestGreen}{$\uparrow\,9.9\%$}}
& 22.6   & \raisebox{-0.35ex}{\scriptsize\textcolor{ForestGreen}{$\uparrow\,26.5\%$}}
& 16.5   & \raisebox{-0.35ex}{\scriptsize\textcolor{ForestGreen}{$\uparrow\,44.4\%$}}
& 12.7   & \raisebox{-0.35ex}{\scriptsize\textcolor{ForestGreen}{$\uparrow\,21.6\%$}}
& 70.8\% & \raisebox{-0.35ex}{\scriptsize\textcolor{ForestGreen}{$\uparrow\,30.1\%$}}
\\
\midrule
\multicolumn{19}{c}{\textbf{Retention Ratio = 0\%}}\\
\midrule
VTW (AAAI'25) \cite{lin2025vtw}
& 879.9  & \raisebox{-0.35ex}{\scriptsize\textcolor{ForestGreen}{$\uparrow\,36.6\%$}}
& 72.9   & \raisebox{-0.35ex}{\scriptsize\textcolor{ForestGreen}{$\uparrow\,15.0\%$}}
& 31.2   & \raisebox{-0.35ex}{\scriptsize\textcolor{ForestGreen}{$\uparrow\,25.4\%$}}
& 42.0   & \raisebox{-0.35ex}{\scriptsize\textcolor{ForestGreen}{$\uparrow\,25.0\%$}}
& 29.2   & \raisebox{-0.35ex}{\scriptsize\textcolor{ForestGreen}{$\uparrow\,21.4\%$}}
& 6.4    & \raisebox{-0.35ex}{\scriptsize\textcolor{ForestGreen}{$\uparrow\,41.0\%$}}
& 4.2    & \raisebox{-0.35ex}{\scriptsize\textcolor{ForestGreen}{$\uparrow\,52.8\%$}}
& 11.6   & \raisebox{-0.35ex}{\scriptsize\textcolor{ForestGreen}{$\uparrow\,25.5\%$}}
& 44.5\% & \raisebox{-0.35ex}{\scriptsize\textcolor{ForestGreen}{$\uparrow\,31.0\%$}}
\\
\bottomrule[1pt]
\end{tabular}}
\caption{\textbf{Comparisons of visual token compression methods in the initial denoising step of LLaDA-V.} 
Each cell reports the evaluation score. 
For the LLaDA-V baseline, bottom-right values indicate its reference inference time (min:sec), 
while for compression methods, the bottom-right values show the relative time change with respect to the baseline 
(\textcolor{ForestGreen}{$\uparrow$~faster / time reduced}, \textcolor{BrickRed}{$\downarrow$~slower / time increased}).
Across different pruning ratios, existing compression methods achieve moderate efficiency gains while causing significant performance drops. Notably, VTW prunes all visual tokens at a specific layer and is therefore excluded from the comparison at other retention ratios.}
\label{tab:lladav}
\end{table*}

\subsection{Computational Cost Analysis}
\label{sec:preliminary_comput_cost}

Following prior studies~\cite{chen2024fastv, wen2025dart, li2025todre}, the floating-point operations (FLOPs) of Transformer-based MLLMs can be formulated as:

\begin{equation}
\mathrm{FLOPs}_{\text{prefilling}}
=
T\times(4n d^{2}+2n^{2}d+2ndm),
\end{equation}
\begin{equation}
\begin{split}
\mathrm{FLOPs}_{\text{decoding}}
&= T\sum_{t=1}^{L}(4 d^{2}+2d(n+t-1)+2d m)\\
&= T(4L d^{2}+2Ld m+dL(2n+L-1)),
\end{split}
\end{equation}
where $T$ denotes the number of Transformer layers; $n$ is the input sequence length; $L$ stands for the output length; $d$ represents the hidden size; $m$ denotes the intermediate dimension of feed-forward networks.

For dMLLMs, since each denoising step performs 
bidirectional interactions among tokens, the total FLOPs can be defined as:

\begin{equation}
\mathrm{FLOPs}_{\text{decoding}}
=
T\sum_{s=1}^{S}\!\bigl(4n_s d^{2}+2n_s^{2}d+2n_s d m\bigr),
\end{equation}
where $S$ is the number of denoising steps, $n_s$ represents the number of tokens actively involved in computation at step $s$, $T$ remains the number of Transformer layers, and the other symbols follow the same definitions as above. 
Note that this formulation only provides an approximation of FLOPs.
The exact count may vary due to internal caching mechanisms and sampling strategies of individual dMLLMs, which determine the actual number of active tokens across different steps and computation stages.

Theoretically, dMLLMs incur heavier computational overhead than MLLMs, as each denoising step performs a prefilling-style computation with bidirectional attention. 
The linear and quadratic terms on the active token number $n_s$ significantly increase the computational cost for multimodal inputs with excessive visual tokens, making it especially beneficial to reduce redundant visual tokens to alleviate this computational burden.
\section{Experiments}
\label{sec:experiments}

\newcommand{\scoreDelta}[3]{%
  \makecell[c]{\begin{tabular}{@{}r@{}l@{}}#1 & \raisebox{-0.35ex}{\scriptsize\textcolor{#2}{#3}}\end{tabular}}%
}

\begin{table*}[!t]
\centering
\resizebox{\linewidth}{!}{%
\begin{tabular}{l|*{8}{r@{}l}|r@{}l}
\toprule[1pt]
\multicolumn{1}{c|}{\textbf{Method}} & \multicolumn{2}{c}{\textbf{MME} \cite{Fu2023mme}} & \multicolumn{2}{c}{\textbf{SQA} \cite{lu2022sqa}} & \multicolumn{2}{c}{\textbf{GQA} \cite{hudson2019gqa}} & \multicolumn{2}{c}{\textbf{POPE} \cite{li2023pope}} & 
\multicolumn{2}{c}{\textbf{MMB} \cite{liu2024mmbench}} & \multicolumn{2}{c}{\textbf{TVQA} \cite{singh2019textvqa}} & \multicolumn{2}{c}{\textbf{CQA} \cite{masry2022chartqa}} & \multicolumn{2}{c}{\textbf{MMMUP} \cite{yue2024mmmupro}} & \multicolumn{2}{c}{\textbf{Avg.}}\\
\midrule

LaViDa-Dream \cite{li2025lavida}
& 1857.0 & \makecell[c]{\\ \textcolor{gray}{5:37}}
& 75.3   & \makecell[c]{\\ \textcolor{gray}{4:28}}
& 56.4   & \makecell[c]{\\ \textcolor{gray}{27:53}}
& 87.0   & \makecell[c]{\\ \textcolor{gray}{20:04}}
& 73.6   & \makecell[c]{\\ \textcolor{gray}{9:31}}
& 42.4   & \makecell[c]{\\ \textcolor{gray}{11:35}}
& 60.4   & \makecell[c]{\\ \textcolor{gray}{6:35}}
& 13.5   & \makecell[c]{\\ \textcolor{gray}{4:22}}
& 100.0\%& \makecell[c]{\\ \textcolor{gray}{90:05}}\\

\midrule
\multicolumn{19}{c}{\textbf{Retention Ratio = 50\%}}\\
\midrule
ToMe (ICLR'23) \cite{bolya2022tome}
& 1638.6 & \raisebox{-0.35ex}{\scriptsize\textcolor{ForestGreen}{$\uparrow\,3.9\%$}} 
& 74.6   & \raisebox{-0.35ex}{\scriptsize\textcolor{ForestGreen}{$\uparrow\,0.4\%$}}
& 53.0   & \raisebox{-0.35ex}{\scriptsize\textcolor{ForestGreen}{$\uparrow\,2.2\%$}}
& 83.7   & \raisebox{-0.35ex}{\scriptsize\textcolor{ForestGreen}{$\uparrow\,3.7\%$}}
& 68.5   & \raisebox{-0.35ex}{\scriptsize\textcolor{ForestGreen}{$\uparrow\,2.5\%$}}
& 31.3   & \raisebox{-0.35ex}{\scriptsize\textcolor{ForestGreen}{$\uparrow\,4.3\%$}}
& 33.9   & \raisebox{-0.35ex}{\scriptsize\textcolor{ForestGreen}{$\uparrow\,4.1\%$}}
& 10.6   & \raisebox{-0.35ex}{\scriptsize\textcolor{ForestGreen}{$\uparrow\,4.2\%$}}
& 84.9\% & \raisebox{-0.35ex}{\scriptsize\textcolor{ForestGreen}{$\uparrow\,3.1\%$}}
\\
FastV (ECCV'24) \cite{chen2024fastv}
& 1771.1 & \raisebox{-0.35ex}{\scriptsize\textcolor{BrickRed}{$\downarrow\,40.9\%$}}
& 73.4   & \raisebox{-0.35ex}{\scriptsize\textcolor{BrickRed}{$\downarrow\,39.2\%$}}
& 52.2   & \raisebox{-0.35ex}{\scriptsize\textcolor{BrickRed}{$\downarrow\,51.2\%$}}
& 84.3   & \raisebox{-0.35ex}{\scriptsize\textcolor{BrickRed}{$\downarrow\,45.5\%$}}
& 69.3   & \raisebox{-0.35ex}{\scriptsize\textcolor{BrickRed}{$\downarrow\,37.1\%$}}
& 30.6   & \raisebox{-0.35ex}{\scriptsize\textcolor{BrickRed}{$\downarrow\,50.4\%$}}
& 22.8   & \raisebox{-0.35ex}{\scriptsize\textcolor{BrickRed}{$\downarrow\,55.4\%$}}
& 12.6   & \raisebox{-0.35ex}{\scriptsize\textcolor{BrickRed}{$\downarrow\,51.5\%$}}
& 85.0\% & \raisebox{-0.35ex}{\scriptsize\textcolor{BrickRed}{$\downarrow\,47.4\%$}}
\\
TRIM (COLING'25) \cite{ruiz2024trim}
& 1589.1 & \raisebox{-0.35ex}{\scriptsize\textcolor{ForestGreen}{$\uparrow\,1.5\%$}}
& 75.1   & \raisebox{-0.35ex}{\scriptsize\textcolor{BrickRed}{$\downarrow\,0.7\%$}}
& 49.6   & \raisebox{-0.35ex}{\scriptsize\textcolor{ForestGreen}{$\uparrow\,0.0\%$}}
& 78.4   & \raisebox{-0.35ex}{\scriptsize\textcolor{ForestGreen}{$\uparrow\,0.8\%$}}
& 69.9   & \raisebox{-0.35ex}{\scriptsize\textcolor{ForestGreen}{$\uparrow\,0.4\%$}}
& 28.2   & \raisebox{-0.35ex}{\scriptsize\textcolor{ForestGreen}{$\uparrow\,2.6\%$}}
& 31.4   & \raisebox{-0.35ex}{\scriptsize\textcolor{ForestGreen}{$\uparrow\,2.8\%$}}
& 11.0   & \raisebox{-0.35ex}{\scriptsize\textcolor{ForestGreen}{$\uparrow\,3.4\%$}}
& 82.3\% & \raisebox{-0.35ex}{\scriptsize\textcolor{ForestGreen}{$\uparrow\,1.0\%$}}
\\
SparseVLM (ICML'25) \cite{zhang2024sparsevlm}
& 1798.9 & \raisebox{-0.35ex}{\scriptsize\textcolor{BrickRed}{$\downarrow\,54.0\%$}}
& 56.2   & \raisebox{-0.35ex}{\scriptsize\textcolor{BrickRed}{$\downarrow\,37.3\%$}}
& 54.5   & \raisebox{-0.35ex}{\scriptsize\textcolor{BrickRed}{$\downarrow\,45.2\%$}}
& 83.9   & \raisebox{-0.35ex}{\scriptsize\textcolor{BrickRed}{$\downarrow\,63.1\%$}}
& 69.7   & \raisebox{-0.35ex}{\scriptsize\textcolor{BrickRed}{$\downarrow\,37.0\%$}}
& 29.3   & \raisebox{-0.35ex}{\scriptsize\textcolor{BrickRed}{$\downarrow\,67.5\%$}}
& 27.1   & \raisebox{-0.35ex}{\scriptsize\textcolor{BrickRed}{$\downarrow\,67.6\%$}}
& 10.8   & \raisebox{-0.35ex}{\scriptsize\textcolor{BrickRed}{$\downarrow\,51.1\%$}}
& 81.6\% & \raisebox{-0.35ex}{\scriptsize\textcolor{BrickRed}{$\downarrow\,53.3\%$}}
\\
DivPrune (CVPR'25) \cite{alvar2025divprune}
& 1683.5 & \raisebox{-0.35ex}{\scriptsize\textcolor{BrickRed}{$\downarrow\,1.2\%$}}
& 74.2   & \raisebox{-0.35ex}{\scriptsize\textcolor{BrickRed}{$\downarrow\,1.5\%$}}
& 52.8   & \raisebox{-0.35ex}{\scriptsize\textcolor{BrickRed}{$\downarrow\,0.7\%$}}
& 85.2   & \raisebox{-0.35ex}{\scriptsize\textcolor{BrickRed}{$\downarrow\,2.5\%$}}
& 66.7   & \raisebox{-0.35ex}{\scriptsize\textcolor{BrickRed}{$\downarrow\,0.2\%$}}
& 32.3   & \raisebox{-0.35ex}{\scriptsize\textcolor{BrickRed}{$\downarrow\,0.6\%$}}
& 30.8   & \raisebox{-0.35ex}{\scriptsize\textcolor{BrickRed}{$\downarrow\,0.5\%$}}
& 13.1   & \raisebox{-0.35ex}{\scriptsize\textcolor{BrickRed}{$\downarrow\,0.8\%$}}
& 87.0\% & \raisebox{-0.35ex}{\scriptsize\textcolor{BrickRed}{$\downarrow\,1.1\%$}}
\\
\midrule
\multicolumn{19}{c}{\textbf{Retention Ratio = 25\%}}\\
\midrule
ToMe (ICLR'23) \cite{bolya2022tome}
& 1606.5 & \raisebox{-0.35ex}{\scriptsize\textcolor{ForestGreen}{$\uparrow\,5.3\%$}}
& 74.3   & \raisebox{-0.35ex}{\scriptsize\textcolor{ForestGreen}{$\uparrow\,1.1\%$}}
& 51.8   & \raisebox{-0.35ex}{\scriptsize\textcolor{ForestGreen}{$\uparrow\,4.1\%$}}
& 81.7   & \raisebox{-0.35ex}{\scriptsize\textcolor{ForestGreen}{$\uparrow\,4.2\%$}}
& 67.4   & \raisebox{-0.35ex}{\scriptsize\textcolor{ForestGreen}{$\uparrow\,3.9\%$}}
& 28.9   & \raisebox{-0.35ex}{\scriptsize\textcolor{ForestGreen}{$\uparrow\,6.2\%$}}
& 29.4   & \raisebox{-0.35ex}{\scriptsize\textcolor{ForestGreen}{$\uparrow\,4.3\%$}}
& 10.6   & \raisebox{-0.35ex}{\scriptsize\textcolor{ForestGreen}{$\uparrow\,5.0\%$}}
& 82.2\% & \raisebox{-0.35ex}{\scriptsize\textcolor{ForestGreen}{$\uparrow\,4.3\%$}}
\\
FastV (ECCV'24) \cite{chen2024fastv}
& 1593.7 & \raisebox{-0.35ex}{\scriptsize\textcolor{BrickRed}{$\downarrow\,30.3\%$}}
& 72.1   & \raisebox{-0.35ex}{\scriptsize\textcolor{BrickRed}{$\downarrow\,31.3\%$}}
& 47.8   & \raisebox{-0.35ex}{\scriptsize\textcolor{BrickRed}{$\downarrow\,28.9\%$}}
& 75.3   & \raisebox{-0.35ex}{\scriptsize\textcolor{BrickRed}{$\downarrow\,30.7\%$}}
& 64.6   & \raisebox{-0.35ex}{\scriptsize\textcolor{BrickRed}{$\downarrow\,26.4\%$}}
& 19.6   & \raisebox{-0.35ex}{\scriptsize\textcolor{BrickRed}{$\downarrow\,35.0\%$}}
& 15.1   & \raisebox{-0.35ex}{\scriptsize\textcolor{BrickRed}{$\downarrow\,40.3\%$}}
& 10.2   & \raisebox{-0.35ex}{\scriptsize\textcolor{BrickRed}{$\downarrow\,33.6\%$}}
& 73.4\% & \raisebox{-0.35ex}{\scriptsize\textcolor{BrickRed}{$\downarrow\,31.1\%$}}
\\
TRIM (COLING'25) \cite{ruiz2024trim}
& 1552.1 & \raisebox{-0.35ex}{\scriptsize\textcolor{ForestGreen}{$\uparrow\,2.4\%$}}
& 74.3   & \raisebox{-0.35ex}{\scriptsize\textcolor{BrickRed}{$\downarrow\,1.9\%$}}
& 46.1   & \raisebox{-0.35ex}{\scriptsize\textcolor{ForestGreen}{$\uparrow\,1.1\%$}}
& 70.8   & \raisebox{-0.35ex}{\scriptsize\textcolor{ForestGreen}{$\uparrow\,3.2\%$}}
& 67.6   & \raisebox{-0.35ex}{\scriptsize\textcolor{ForestGreen}{$\uparrow\,1.6\%$}}
& 24.1   & \raisebox{-0.35ex}{\scriptsize\textcolor{ForestGreen}{$\uparrow\,3.2\%$}}
& 27.3   & \raisebox{-0.35ex}{\scriptsize\textcolor{ForestGreen}{$\uparrow\,3.8\%$}}
& 10.9   & \raisebox{-0.35ex}{\scriptsize\textcolor{ForestGreen}{$\uparrow\,5.0\%$}}
& 77.5\% & \raisebox{-0.35ex}{\scriptsize\textcolor{ForestGreen}{$\uparrow\,2.2\%$}}
\\
SparseVLM (ICML'25) \cite{zhang2024sparsevlm}
& 1582.7 & \raisebox{-0.35ex}{\scriptsize\textcolor{BrickRed}{$\downarrow\,41.5\%$}}
& 52.3   & \raisebox{-0.35ex}{\scriptsize\textcolor{BrickRed}{$\downarrow\,31.3\%$}}
& 49.4   & \raisebox{-0.35ex}{\scriptsize\textcolor{BrickRed}{$\downarrow\,31.4\%$}}
& 76.2   & \raisebox{-0.35ex}{\scriptsize\textcolor{BrickRed}{$\downarrow\,48.9\%$}}
& 66.2   & \raisebox{-0.35ex}{\scriptsize\textcolor{BrickRed}{$\downarrow\,32.9\%$}}
& 19.9   & \raisebox{-0.35ex}{\scriptsize\textcolor{BrickRed}{$\downarrow\,54.7\%$}}
& 14.6   & \raisebox{-0.35ex}{\scriptsize\textcolor{BrickRed}{$\downarrow\,54.4\%$}}
& 10.5   & \raisebox{-0.35ex}{\scriptsize\textcolor{BrickRed}{$\downarrow\,42.7\%$}}
& 71.1\% & \raisebox{-0.35ex}{\scriptsize\textcolor{BrickRed}{$\downarrow\,41.3\%$}}
\\
DivPrune (CVPR'25) \cite{alvar2025divprune}
& 1442.6 & \raisebox{-0.35ex}{\scriptsize\textcolor{ForestGreen}{$\uparrow\,1.5\%$}}
& 73.8   & \raisebox{-0.35ex}{\scriptsize\textcolor{BrickRed}{$\downarrow\,1.1\%$}}
& 48.8   & \raisebox{-0.35ex}{\scriptsize\textcolor{ForestGreen}{$\uparrow\,0.5\%$}}
& 77.4   & \raisebox{-0.35ex}{\scriptsize\textcolor{ForestGreen}{$\uparrow\,0.2\%$}}
& 59.1   & \raisebox{-0.35ex}{\scriptsize\textcolor{ForestGreen}{$\uparrow\,2.5\%$}}
& 23.9   & \raisebox{-0.35ex}{\scriptsize\textcolor{ForestGreen}{$\uparrow\,1.3\%$}}
& 20.7   & \raisebox{-0.35ex}{\scriptsize\textcolor{ForestGreen}{$\uparrow\,1.8\%$}}
& 12.2   & \raisebox{-0.35ex}{\scriptsize\textcolor{ForestGreen}{$\uparrow\,2.7\%$}}
& 76.6\% & \raisebox{-0.35ex}{\scriptsize\textcolor{ForestGreen}{$\uparrow\,0.9\%$}}
\\
\midrule
\multicolumn{19}{c}{\textbf{Retention Ratio = 10\%}}\\
\midrule
ToMe (ICLR'23) \cite{bolya2022tome}
& 1558.9 & \raisebox{-0.35ex}{\scriptsize\textcolor{ForestGreen}{$\uparrow\,6.5\%$}}
& 74.1   & \raisebox{-0.35ex}{\scriptsize\textcolor{ForestGreen}{$\uparrow\,3.7\%$}}
& 49.3   & \raisebox{-0.35ex}{\scriptsize\textcolor{ForestGreen}{$\uparrow\,4.8\%$}}
& 76.3   & \raisebox{-0.35ex}{\scriptsize\textcolor{ForestGreen}{$\uparrow\,5.3\%$}}
& 62.8   & \raisebox{-0.35ex}{\scriptsize\textcolor{ForestGreen}{$\uparrow\,4.0\%$}}
& 21.4   & \raisebox{-0.35ex}{\scriptsize\textcolor{ForestGreen}{$\uparrow\,7.2\%$}}
& 22.4   & \raisebox{-0.35ex}{\scriptsize\textcolor{ForestGreen}{$\uparrow\,6.8\%$}}
& 11.1   & \raisebox{-0.35ex}{\scriptsize\textcolor{ForestGreen}{$\uparrow\,4.6\%$}}
& 76.6\% & \raisebox{-0.35ex}{\scriptsize\textcolor{ForestGreen}{$\uparrow\,5.3\%$}}
\\
FastV (ECCV'24) \cite{chen2024fastv}
& 1374.1 & \raisebox{-0.35ex}{\scriptsize\textcolor{BrickRed}{$\downarrow\,19.9\%$}}
& 71.2   & \raisebox{-0.35ex}{\scriptsize\textcolor{BrickRed}{$\downarrow\,15.3\%$}}
& 42.8   & \raisebox{-0.35ex}{\scriptsize\textcolor{BrickRed}{$\downarrow\,21.9\%$}}
& 61.2   & \raisebox{-0.35ex}{\scriptsize\textcolor{BrickRed}{$\downarrow\,22.8\%$}}
& 56.8   & \raisebox{-0.35ex}{\scriptsize\textcolor{BrickRed}{$\downarrow\,17.7\%$}}
& 12.6   & \raisebox{-0.35ex}{\scriptsize\textcolor{BrickRed}{$\downarrow\,23.2\%$}}
& 12.9   & \raisebox{-0.35ex}{\scriptsize\textcolor{BrickRed}{$\downarrow\,26.6\%$}}
& 10.8   & \raisebox{-0.35ex}{\scriptsize\textcolor{BrickRed}{$\downarrow\,21.4\%$}}
& 65.4\% & \raisebox{-0.35ex}{\scriptsize\textcolor{BrickRed}{$\downarrow\,21.7\%$}}
\\
TRIM (COLING'25) \cite{ruiz2024trim}
& 1385.4 & \raisebox{-0.35ex}{\scriptsize\textcolor{ForestGreen}{$\uparrow\,3.0\%$}}
& 72.8   & \raisebox{-0.35ex}{\scriptsize\textcolor{BrickRed}{$\downarrow\,1.5\%$}}
& 43.2   & \raisebox{-0.35ex}{\scriptsize\textcolor{ForestGreen}{$\uparrow\,1.9\%$}}
& 62.4   & \raisebox{-0.35ex}{\scriptsize\textcolor{ForestGreen}{$\uparrow\,3.9\%$}}
& 61.1   & \raisebox{-0.35ex}{\scriptsize\textcolor{ForestGreen}{$\uparrow\,1.8\%$}}
& 19.6   & \raisebox{-0.35ex}{\scriptsize\textcolor{ForestGreen}{$\uparrow\,3.6\%$}}
& 17.4   & \raisebox{-0.35ex}{\scriptsize\textcolor{ForestGreen}{$\uparrow\,3.5\%$}}
& 11.3   & \raisebox{-0.35ex}{\scriptsize\textcolor{ForestGreen}{$\uparrow\,8.0\%$}}
& 70.1\% & \raisebox{-0.35ex}{\scriptsize\textcolor{ForestGreen}{$\uparrow\,2.8\%$}}
\\
SparseVLM (ICML'25) \cite{zhang2024sparsevlm}
& 1531.4 & \raisebox{-0.35ex}{\scriptsize\textcolor{BrickRed}{$\downarrow\,35.6\%$}}
& 51.4   & \raisebox{-0.35ex}{\scriptsize\textcolor{BrickRed}{$\downarrow\,19.0\%$}}
& 45.2   & \raisebox{-0.35ex}{\scriptsize\textcolor{BrickRed}{$\downarrow\,21.9\%$}}
& 67.4   & \raisebox{-0.35ex}{\scriptsize\textcolor{BrickRed}{$\downarrow\,44.8\%$}}
& 60.8   & \raisebox{-0.35ex}{\scriptsize\textcolor{BrickRed}{$\downarrow\,27.8\%$}}
& 15.0   & \raisebox{-0.35ex}{\scriptsize\textcolor{BrickRed}{$\downarrow\,50.2\%$}}
& 11.6   & \raisebox{-0.35ex}{\scriptsize\textcolor{BrickRed}{$\downarrow\,47.6\%$}}
& 10.1   & \raisebox{-0.35ex}{\scriptsize\textcolor{BrickRed}{$\downarrow\,43.1\%$}}
& 65.0\% & \raisebox{-0.35ex}{\scriptsize\textcolor{BrickRed}{$\downarrow\,34.9\%$}}
\\
DivPrune (CVPR'25) \cite{alvar2025divprune}
& 1230.9 & \raisebox{-0.35ex}{\scriptsize\textcolor{ForestGreen}{$\uparrow\,3.6\%$}}
& 72.3   & \raisebox{-0.35ex}{\scriptsize\textcolor{ForestGreen}{$\uparrow\,0.4\%$}}
& 43.7   & \raisebox{-0.35ex}{\scriptsize\textcolor{ForestGreen}{$\uparrow\,3.8\%$}}
& 64.0   & \raisebox{-0.35ex}{\scriptsize\textcolor{ForestGreen}{$\uparrow\,2.9\%$}}
& 45.5   & \raisebox{-0.35ex}{\scriptsize\textcolor{ForestGreen}{$\uparrow\,3.5\%$}}
& 16.3   & \raisebox{-0.35ex}{\scriptsize\textcolor{ForestGreen}{$\uparrow\,3.2\%$}}
& 15.3   & \raisebox{-0.35ex}{\scriptsize\textcolor{ForestGreen}{$\uparrow\,3.3\%$}}
& 10.8   & \raisebox{-0.35ex}{\scriptsize\textcolor{ForestGreen}{$\uparrow\,3.8\%$}}
& 64.9\% & \raisebox{-0.35ex}{\scriptsize\textcolor{ForestGreen}{$\uparrow\,3.3\%$}}
\\
\midrule
\multicolumn{19}{c}{\textbf{Retention Ratio = 0\%}}\\
\midrule
VTW (AAAI'25) \cite{lin2025vtw}
& 1733.3 & \raisebox{-0.35ex}{\scriptsize\textcolor{ForestGreen}{$\uparrow\,0.9\%$}}
& 74.9   & \raisebox{-0.35ex}{\scriptsize\textcolor{ForestGreen}{$\uparrow\,2.6\%$}}
& 43.4   & \raisebox{-0.35ex}{\scriptsize\textcolor{ForestGreen}{$\uparrow\,0.4\%$}}
& 65.2   & \raisebox{-0.35ex}{\scriptsize\textcolor{ForestGreen}{$\uparrow\,0.5\%$}}
& 70.3   & \raisebox{-0.35ex}{\scriptsize\textcolor{ForestGreen}{$\uparrow\,1.6\%$}}
& 13.5   & \raisebox{-0.35ex}{\scriptsize\textcolor{ForestGreen}{$\uparrow\,2.2\%$}}
& 13.2   & \raisebox{-0.35ex}{\scriptsize\textcolor{ForestGreen}{$\uparrow\,0.3\%$}}
& 11.4   & \raisebox{-0.35ex}{\scriptsize\textcolor{ForestGreen}{$\uparrow\,1.1\%$}}
& 72.3\% & \raisebox{-0.35ex}{\scriptsize\textcolor{ForestGreen}{$\uparrow\,0.9\%$}}
\\
\bottomrule[1pt]
\end{tabular}}
\caption{\textbf{Comparison of visual token compression methods in the initial denoising step of LaViDa-Dream.} 
Each cell reports the evaluation score. 
For the LaViDa-Dream baseline, bottom-right values indicate its reference inference time (min:sec), 
while for compression methods, the bottom-right values show the relative time change with respect to the baseline 
(\textcolor{ForestGreen}{$\uparrow$~faster / time reduced}, \textcolor{BrickRed}{$\downarrow$~slower / time increased}).
Across different pruning ratios, existing compression methods achieve only negligible efficiency gains while causing significant performance drops. Notably, FastV and SparseVLM are even slower than the baseline due to their incompatibility with efficient attention operators \cite{dao2022flashattention}.}
\label{tab:lavida}
\end{table*}

We conduct extensive experiments to investigate whether visual token redundancy exists in dMLLMs and how visual token pruning affects the inference efficiency and accuracy of dMLLMs. Specifically, we study dMLLMs by adapting representative pruning methods that have demonstrated impressive performance for autoregressive MLLMs.

\subsection{Experimental Settings}
\label{subsec:exp_setting}

\paragraph{Backbones and Benchmarks.}
We conduct experiments with LLaDA-V~\cite{you2025lladav} and LaViDa-Dream~\cite{li2025lavida}, two representative dMLLM backbones that are trained via \emph{from-scratch} and \emph{AR-to-diffusion}, respectively. In addition, we evaluate both backbones over twelve widely adopted benchmarks, including ten for image understanding and two for video understanding. All experiments are conducted on 8 × A800 GPUs. More details of the backbones, benchmarks and compression methods are provided in the Appendix. \looseness=-1

\paragraph{Token Compression Methods.}


As described in \Cref{sec:related_work_vtc}, most existing visual token pruning methods adopt four underlying pruning mechanisms, namely, similarity-based pruning, attention-based pruning, query-based pruning, and transformation-based pruning. Unlike the other three visual token pruning approaches, transformer-based pruning is a built-in design in both MLLMs and dMLLMs which operates at a fixed rate and is non-adjustable at inference time \cite{wang2024qwen2, bai2025qwen25vl, li2024llavaonevision}. We therefore exclude the transformer-based pruning and focus on the other three pruning approaches in our experiments. Specifically, we select two representative methods for each of the three pruning approaches, including ToMe~\cite{bolya2022tome} and DivPrune~\cite{alvar2025divprune} for similarity-based pruning, FastV~\cite{chen2024fastv}, VTW~\cite{lin2025vtw} for attention-based pruning, and SparseVLM~\cite{zhang2024sparsevlm} and TRIM~\cite{ruiz2024trim} for query-based pruning.

\subsection{Short-answer Tasks: No Visual Redundancy}
\label{subsec:short-answer}

We first conduct experiments on both LLaDA-V and LaViDa-Dream (see Table \ref{tab:lladav} and Table \ref{tab:lavida}) under varying token retention ratios during the initial denoising step.
For LLaDA-V, various token compression methods exhibit a consistent pattern: models either show slight degradation in performance with only marginal efficiency gains at a retention ratio of 50\%, or achieve minimal acceleration at the cost of significant performance degradation when the retention ratio is below 50\%.
For LaViDa-Dream, existing compression methods perform even worse—exhibiting severe performance degradation while offering negligible speed-ups across all retention ratios; some are even slower than the baseline due to their incompatibility with efficient attention operators~\cite{dao2022flashattention}.
These findings indicate that for both architectures, the trade-off between efficiency and performance is largely ineffective.

Given that each denoising step in dMLLMs performs a similar bidirectional computation, it is natural to further explore applying these compression methods to later denoising steps for potential acceleration. 
As shown in Table~\ref{tab:decoding-steps} (short-answer tasks), compression is applied to different portions of denoising steps (the percentage indicates the fraction of total denoising steps to which pruning is applied).
For LLaDA-V, pruning visual tokens provides only marginal efficiency gains while causing a noticeable performance drop, whereas for LaViDa-Dream, pruning yields almost no acceleration effect.
The results remain consistent with those from applying pruning in the first denoising step: pruning visual tokens during intermediate steps still yields a poor trade-off between performance and efficiency across both architectures. \looseness=-1

Overall, these findings suggest that existing compression techniques\textemdash{}though based on diverse underlying mechanisms\textemdash{}fail to effectively remove redundant visual tokens, implying that each token contributes meaningfully to the reasoning process in short-answer tasks.

\begin{table*}[t]
\centering
\small
\setlength{\tabcolsep}{4.5pt}
\renewcommand{\arraystretch}{1.16}
\resizebox{\textwidth}{!}{%
\begin{tabular}{l|*{3}{r@{}l}|*{3}{r@{}l}|*{3}{r@{}l}|*{3}{r@{}l}}
\toprule[1pt]
& \multicolumn{6}{c|}{\textbf{Short-answer Tasks}} 
& \multicolumn{18}{c}{\textbf{Long-answer Tasks}} \\
\cmidrule(lr){2-7}\cmidrule(lr){8-25}
\multicolumn{1}{c|}{\textbf{\large Method}}
& \multicolumn{2}{c}{\textbf{MME} \cite{Fu2023mme}} & \multicolumn{2}{c}{\textbf{POPE} \cite{li2023pope}} & \multicolumn{2}{c|}{\textbf{VMME} \cite{fu2024videomme}}
& \multicolumn{6}{c|}{\textbf{IVQA} \cite{mathew2022infographicvqa}} 
& \multicolumn{6}{c|}{\textbf{DVQA} \cite{mathew2021docvqa}} 
& \multicolumn{6}{c}{\textbf{VDC} \cite{lmmslab2024videodetailcaption}} \\
\cmidrule(lr){2-7}\cmidrule(lr){8-13}\cmidrule(lr){14-19}\cmidrule(lr){20-25}
& \multicolumn{2}{c}{50\%} & \multicolumn{2}{c}{50\%} & \multicolumn{2}{c|}{50\%} 
& \multicolumn{2}{c}{25\%} & \multicolumn{2}{c}{50\%} & \multicolumn{2}{c|}{75\%}
& \multicolumn{2}{c}{25\%} & \multicolumn{2}{c}{50\%} & \multicolumn{2}{c|}{75\%}
& \multicolumn{2}{c}{25\%} & \multicolumn{2}{c}{50\%} & \multicolumn{2}{c}{75\%} \\
\midrule
\multicolumn{25}{c}{\textbf{From-scratch dMLLM}}\\
\midrule
LLaDA-V \cite{you2025lladav}
& 1994.4 & \makecell[c]{\\ \textcolor{gray}{5:47}}
& 87.3   & \makecell[c]{\\ \textcolor{gray}{15:45}}
& 56.0   & \makecell[c]{\\ \textcolor{gray}{22:02}}
& \multicolumn{6}{c|}{\makecell[c]{66.2\raisebox{-0.35ex}{\scriptsize\textcolor{gray}{\,42:03}}}} 
& \multicolumn{6}{c|}{\makecell[c]{83.9\raisebox{-0.35ex}{\scriptsize\textcolor{gray}{\,1:22:52}}}} 
& \multicolumn{6}{c}{\makecell[c]{2.8\raisebox{-0.35ex}{\scriptsize\textcolor{gray}{\,56:16}}}} \\
\midrule
FastV \cite{chen2024fastv}
& 1377.0 & \raisebox{-0.35ex}{\scriptsize\textcolor{BrickRed}{$\downarrow\,27.1\%$}}
& 73.8   & \raisebox{-0.35ex}{\scriptsize\textcolor{BrickRed}{$\downarrow\,36.8\%$}}
& 46.3   & \raisebox{-0.35ex}{\scriptsize\textcolor{BrickRed}{$\downarrow\,52.2\%$}}
& 45.9   & \raisebox{-0.35ex}{\scriptsize\textcolor{ForestGreen}{$\uparrow\,14.0\%$}}
& 64.0   & \raisebox{-0.35ex}{\scriptsize\textcolor{BrickRed}{$\downarrow\,31.3\%$}}
& 65.3   & \raisebox{-0.35ex}{\scriptsize\textcolor{BrickRed}{$\downarrow\,75.5\%$}}
& 58.2   & \raisebox{-0.35ex}{\scriptsize\textcolor{ForestGreen}{$\uparrow\,14.8\%$}}
& 81.2   & \raisebox{-0.35ex}{\scriptsize\textcolor{BrickRed}{$\downarrow\,26.9\%$}}
& 83.0   & \raisebox{-0.35ex}{\scriptsize\textcolor{BrickRed}{$\downarrow\,70.4\%$}}
& 2.2    & \raisebox{-0.35ex}{\scriptsize\textcolor{BrickRed}{$\downarrow\,13.5\%$}}
& 2.6    & \raisebox{-0.35ex}{\scriptsize\textcolor{BrickRed}{$\downarrow\,84.4\%$}}
& 2.7    & \raisebox{-0.35ex}{\scriptsize\textcolor{BrickRed}{$\downarrow\,156.9\%$}}\\
VTW \cite{lin2025vtw}
& 1213.9 & \raisebox{-0.35ex}{\scriptsize\textcolor{ForestGreen}{$\uparrow\,12.1\%$}}
& 55.4   & \raisebox{-0.35ex}{\scriptsize\textcolor{ForestGreen}{$\uparrow\,0.7\%$}}
& 46.2   & \raisebox{-0.35ex}{\scriptsize\textcolor{ForestGreen}{$\uparrow\,10.9\%$}}
& 45.3   & \raisebox{-0.35ex}{\scriptsize\textcolor{ForestGreen}{$\uparrow\,64.4\%$}}
& 63.5   & \raisebox{-0.35ex}{\scriptsize\textcolor{ForestGreen}{$\uparrow\,43.8\%$}}
& 65.2   & \raisebox{-0.35ex}{\scriptsize\textcolor{ForestGreen}{$\uparrow\,23.7\%$}}
& 56.5   & \raisebox{-0.35ex}{\scriptsize\textcolor{ForestGreen}{$\uparrow\,63.2\%$}}
& 80.9   & \raisebox{-0.35ex}{\scriptsize\textcolor{ForestGreen}{$\uparrow\,43.2\%$}}
& 83.0   & \raisebox{-0.35ex}{\scriptsize\textcolor{ForestGreen}{$\uparrow\,23.5\%$}}
& 1.9    & \raisebox{-0.35ex}{\scriptsize\textcolor{ForestGreen}{$\uparrow\,69.4\%$}}
& 2.4    & \raisebox{-0.35ex}{\scriptsize\textcolor{ForestGreen}{$\uparrow\,46.5\%$}}
& 2.7    & \raisebox{-0.35ex}{\scriptsize\textcolor{ForestGreen}{$\uparrow\,23.4\%$}}\\
SparseVLM \cite{zhang2024sparsevlm}
& 1467.0 & \raisebox{-0.35ex}{\scriptsize\textcolor{BrickRed}{$\downarrow\,40.1\%$}}
& 76.8   & \raisebox{-0.35ex}{\scriptsize\textcolor{BrickRed}{$\downarrow\,49.3\%$}}
& 49.1   & \raisebox{-0.35ex}{\scriptsize\textcolor{BrickRed}{$\downarrow\,71.7\%$}}
& 46.5   & \raisebox{-0.35ex}{\scriptsize\textcolor{ForestGreen}{$\uparrow\,12.5\%$}}
& 64.0   & \raisebox{-0.35ex}{\scriptsize\textcolor{BrickRed}{$\downarrow\,34.7\%$}}
& 65.5   & \raisebox{-0.35ex}{\scriptsize\textcolor{BrickRed}{$\downarrow\,75.9\%$}}
& 60.0   & \raisebox{-0.35ex}{\scriptsize\textcolor{ForestGreen}{$\uparrow\,13.1\%$}}
& 81.5   & \raisebox{-0.35ex}{\scriptsize\textcolor{BrickRed}{$\downarrow\,30.7\%$}}
& 83.1   & \raisebox{-0.35ex}{\scriptsize\textcolor{BrickRed}{$\downarrow\,71.7\%$}}
& 2.3    & \raisebox{-0.35ex}{\scriptsize\textcolor{BrickRed}{$\downarrow\,13.9\%$}}
& 2.6    & \raisebox{-0.35ex}{\scriptsize\textcolor{BrickRed}{$\downarrow\,86.7\%$}}
& 2.7    & \raisebox{-0.35ex}{\scriptsize\textcolor{BrickRed}{$\downarrow\,160.7\%$}}\\
DivPrune \cite{alvar2025divprune}
& 1807.0 & \raisebox{-0.35ex}{\scriptsize\textcolor{ForestGreen}{$\uparrow\,11.5\%$}}
& 86.4   & \raisebox{-0.35ex}{\scriptsize\textcolor{BrickRed}{$\downarrow\,2.0\%$}}
& 58.0   & \raisebox{-0.35ex}{\scriptsize\textcolor{ForestGreen}{$\uparrow\,7.7\%$}}
& 61.7   & \raisebox{-0.35ex}{\scriptsize\textcolor{ForestGreen}{$\uparrow\,47.1\%$}}
& 65.0   & \raisebox{-0.35ex}{\scriptsize\textcolor{ForestGreen}{$\uparrow\,30.5\%$}}
& 65.8   & \raisebox{-0.35ex}{\scriptsize\textcolor{ForestGreen}{$\uparrow\,14.0\%$}}
& 78.9   & \raisebox{-0.35ex}{\scriptsize\textcolor{ForestGreen}{$\uparrow\,46.4\%$}}
& 83.1   & \raisebox{-0.35ex}{\scriptsize\textcolor{ForestGreen}{$\uparrow\,30.7\%$}}
& 83.5   & \raisebox{-0.35ex}{\scriptsize\textcolor{ForestGreen}{$\uparrow\,13.7\%$}}
& 2.7    & \raisebox{-0.35ex}{\scriptsize\textcolor{ForestGreen}{$\uparrow\,53.2\%$}}
& 2.8    & \raisebox{-0.35ex}{\scriptsize\textcolor{ForestGreen}{$\uparrow\,35.1\%$}}
& 2.8    & \raisebox{-0.35ex}{\scriptsize\textcolor{ForestGreen}{$\uparrow\,17.0\%$}}\\
\midrule
\multicolumn{25}{c}{\textbf{AR-to-diffusion dMLLM}}\\
\midrule
LaViDa-Dream \cite{li2025lavida} 
& 1857.0 & \makecell[c]{\\ \textcolor{gray}{5:37}}
& 87.0   & \makecell[c]{\\ \textcolor{gray}{20:04}}
& \multicolumn{2}{c|}{\makecell[c]{--\\ \textcolor{gray}{--}}}
& \multicolumn{6}{c|}{\makecell[c]{35.5\raisebox{-0.35ex}{\scriptsize\textcolor{gray}{\,8:53}}}} 
& \multicolumn{6}{c|}{\makecell[c]{55.1\raisebox{-0.35ex}{\scriptsize\textcolor{gray}{\,17:07}}}} 
& \multicolumn{2}{c}{\makecell[c]{--\\ \textcolor{gray}{--}}}
& \multicolumn{2}{c}{\makecell[c]{--\\ \textcolor{gray}{--}}}
& \multicolumn{2}{c}{\makecell[c]{--\\ \textcolor{gray}{--}}}\\
\midrule
VTW \cite{lin2025vtw}
& 1835.4 & \raisebox{-0.35ex}{\scriptsize\textcolor{BrickRed}{$\downarrow\,0.9\%$}}
& 86.7   & \raisebox{-0.35ex}{\scriptsize\textcolor{BrickRed}{$\downarrow\,0.8\%$}}
& \multicolumn{2}{c|}{\makecell[c]{--\\ --}}
& 17.7   & \raisebox{-0.35ex}{\scriptsize\textcolor{BrickRed}{$\downarrow\,2.8\%$}}
& 17.7   & \raisebox{-0.35ex}{\scriptsize\textcolor{BrickRed}{$\downarrow\,1.1\%$}}
& 18.3   & \raisebox{-0.35ex}{\scriptsize\textcolor{BrickRed}{$\downarrow\,1.3\%$}}
& 24.0   & \raisebox{-0.35ex}{\scriptsize\textcolor{BrickRed}{$\downarrow\,3.8\%$}}
& 24.1   & \raisebox{-0.35ex}{\scriptsize\textcolor{BrickRed}{$\downarrow\,1.4\%$}}
& 28.3   & \raisebox{-0.35ex}{\scriptsize\textcolor{BrickRed}{$\downarrow\,3.0\%$}}
& \multicolumn{2}{c}{\makecell[c]{--\\ --}}
& \multicolumn{2}{c}{\makecell[c]{--\\ --}}
& \multicolumn{2}{c}{\makecell[c]{--\\ --}}\\
DivPrune \cite{alvar2025divprune}
& 1857.0 & \raisebox{-0.35ex}{\scriptsize\textcolor{BrickRed}{$\downarrow\,2.4\%$}}
& 86.5   & \raisebox{-0.35ex}{\scriptsize\textcolor{BrickRed}{$\downarrow\,2.2\%$}}
& \multicolumn{2}{c|}{\makecell[c]{--\\ --}}
& 15.9   & \raisebox{-0.35ex}{\scriptsize\textcolor{BrickRed}{$\downarrow\,0.6\%$}}
& 15.9   & \raisebox{-0.35ex}{\scriptsize\textcolor{BrickRed}{$\downarrow\,1.1\%$}}
& 16.1   & \raisebox{-0.35ex}{\scriptsize\textcolor{BrickRed}{$\downarrow\,1.9\%$}}
& 23.7   & \raisebox{-0.35ex}{\scriptsize\textcolor{BrickRed}{$\downarrow\,0.3\%$}}
& 23.9   & \raisebox{-0.35ex}{\scriptsize\textcolor{BrickRed}{$\downarrow\,1.9\%$}}
& 25.6   & \raisebox{-0.35ex}{\scriptsize\textcolor{BrickRed}{$\downarrow\,2.4\%$}}
& \multicolumn{2}{c}{\makecell[c]{--\\ --}}
& \multicolumn{2}{c}{\makecell[c]{--\\ --}}
& \multicolumn{2}{c}{\makecell[c]{--\\ --}}\\
\bottomrule[1pt]
\end{tabular}}
\caption{\textbf{Performance comparison when different token compression methods are applied at various denoising steps under a 25\% retention ratio.}
The percentage indicates the fraction of total denoising steps to which pruning is applied; for short-answer tasks, the small number of steps permits only 50\% as a valid ratio.
For each compression method, the top value indicates the evaluation score, and the bottom-right value shows the relative change in inference time compared with the model’s baseline
(\textcolor{ForestGreen}{$\uparrow$ faster / time reduced}, \textcolor{BrickRed}{$\downarrow$ slower / time increased}).
Baseline bottom-right values show the reference inference time (min:sec).
Note that LaViDa is incapable of video understanding\protect\footnotemark{}, and is thus excluded from video benchmarks.
In short-answer tasks, existing compression methods for both LLaDA-V and LaViDa-Dream yield inefficient trade-offs between efficiency and performance.
In long-answer tasks, these methods remain ineffective on LaViDa-Dream, whereas on LLaDA-V some of them (e.g., DivPrune and VTW) deliver notable efficiency gains with nearly lossless performance when pruning begins at middle or later denoising steps.}
\label{tab:decoding-steps}
\end{table*}
\footnotetext{\url{https://github.com/jacklishufan/LaViDa/issues/27}}

\subsection{Long-answer Tasks: Redundancy Diverges across Backbones}
\label{subsec:long-answer}

We further extend our experiments to long-answer tasks. Notably, two backbones exhibit divergent behaviors (see \Cref{tab:decoding-steps}): in LLaDA-V, visual redundancy gradually emerges as pruning is applied to the middle and later denoising steps, whereas in LaViDa-Dream, it remains largely absent. 
For instance, in LLaDA-V, DivPrune~\cite{alvar2025divprune} prunes 75\% of visual tokens in the middle decoding steps, yielding a 1.44$\times$ speed-up on InfoVQA and DocVQA ($-$30.5\% and $-$30.7\% in time) with almost no performance degradation (65.0 vs. 66.2; 83.1 vs. 83.9).
In contrast, in LaViDa-Dream, applying various compression techniques to different portions of denoising steps yields consistent yet suboptimal results, suffering from substantial performance drops (nearly 50\%) alongside reduced efficiency. 

These findings reveal that visual redundancy manifests differently between from-scratch and AR-to-diffusion dMLLMs: in from-scratch dMLLMs, redundancy grows with longer generation length and more steps, whereas it remains negligible in AR-to-diffusion models.
Unlike MLLMs, where visual redundancy consistently appears across both short- and long-answer tasks, dMLLMs exhibit more complex behavior: the emergence of redundancy depends not only on the modeling paradigm—from-scratch diffusion training versus AR-to-diffusion adaptation—but also on the task type, as short- and long-answer tasks may yield different redundancy patterns. 
This intriguing contrast motivates a deeper analysis in the following section.

\section{Understanding Visual Token Redundancy in dMLLMs}
\label{sec:analysis}

\begin{figure*}[t]
  \centering
  \includegraphics[width=\textwidth]{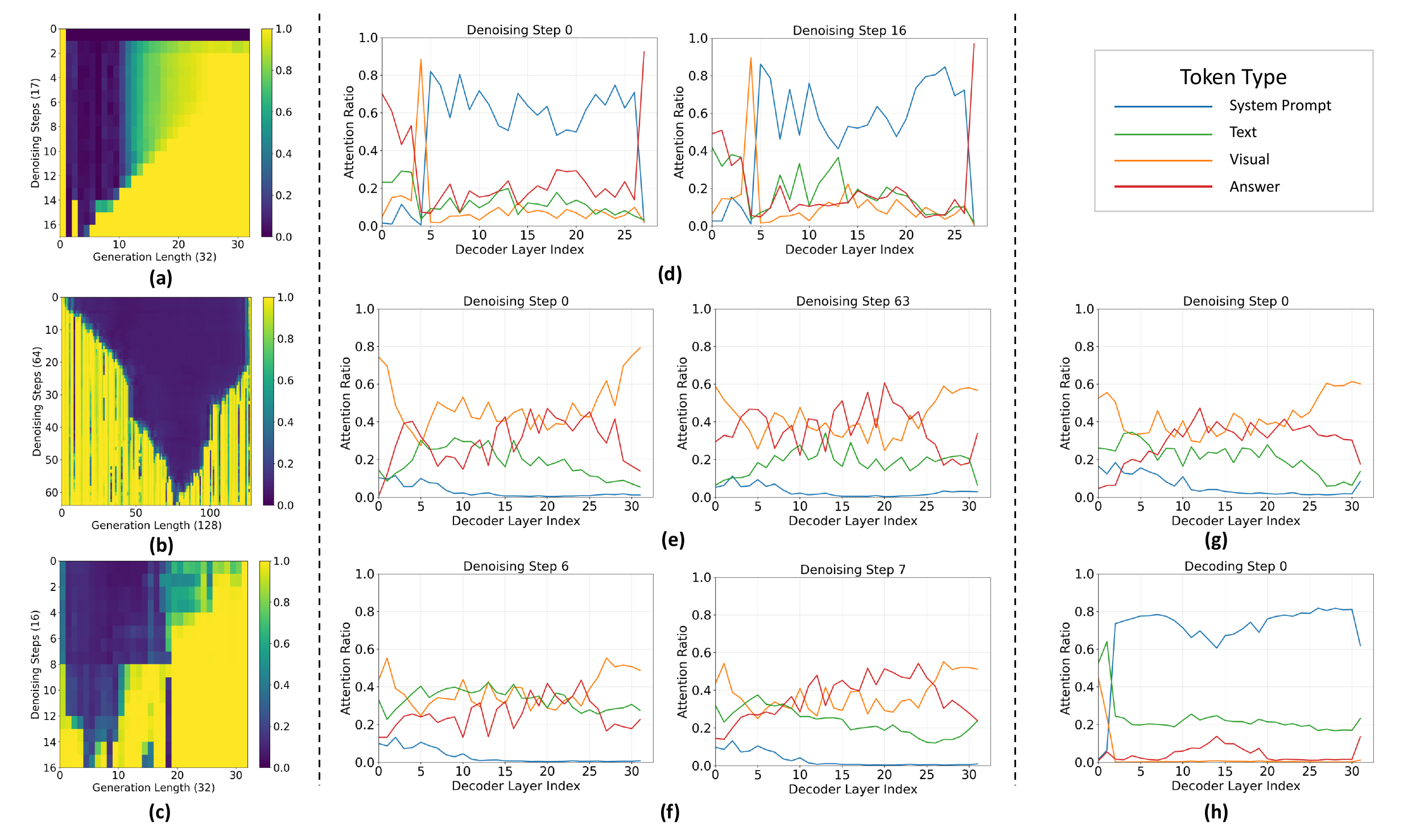}
  \caption{\textbf{Visualization of the fraction of attention from answer tokens to each token type across layers, and of logit dynamics across denoising steps.} 
  (a, d) show the heatmaps and attention ratio variations of LaViDa-Dream \cite{li2025lavida} (representing AR-to-diffusion dMLLMs) across both short- and long-answer tasks.
  (b, e) correspond to LLaDA-V \cite{you2025lladav} (representing from-scratch dMLLMs) on paragraph-level long-answer tasks (Video Detail Description \cite{lmmslab2024videodetailcaption}).
  (c, f) present LLaDA-V on sentence-level long-answer tasks (InfoVQA \cite{mathew2022infographicvqa} and DocVQA \cite{mathew2021docvqa}).
  (g) shows the attention ratio trends of LLaDA-V on short-answer tasks, and (h) depicts those of MLLMs on both short- and long-answer tasks.
  We observe three key patterns:
  (1) Compared with MLLMs, both dMLLM variants exhibit stronger reliance on visual tokens, leading to a more pronounced performance drop when visual tokens are pruned.
  (2) The self-attention intensity among answer tokens progressively increases from MLLMs to LaViDa-Dream and LLaDA-V, endowing dMLLMs, especially LLaDA-V, with a stronger capacity to recover lost information through bidirectional contextual refinement.
  (3) Steps 7–8 in (c) and (f) reveal a consistent pattern where a rise in logits follows a preceding surge in answer-token self-attention, suggesting that stronger self-attention facilitates information restoration during denoising.
  }
  \label{fig:observation}
\end{figure*}

In this section, we analyze the underlying causes of the divergent behaviors observed in from-scratch trained and AR-to-diffusion adapted dMLLMs across short-answer and long-answer tasks. These behaviors can be interpreted through two key factors: the information loss introduced by visual token pruning and the model’s ability to restore this lost information during later denoising. Viewed from these perspectives, visual redundancy persists in dMLLMs, offering new directions for future acceleration strategies.

\subsection{Information Loss from Visual Token Pruning}
\label{subsec:impact_of_pruning}

\textbf{Takeaway: }\textit{Pruning visual tokens leads to more severe information loss in dMLLMs than in MLLMs, as both from-scratch and AR-to-diffusion dMLLMs exhibit a much stronger reliance on visual information compared with MLLMs.}

As shown in \Cref{fig:observation} (h), answer tokens in MLLMs exhibit near-zero attention to visual tokens (the orange line) throughout the entire decoding process. This phenomenon remains highly consistent across different model scales and both short- and long-answer tasks, despite the large number of visual tokens present. This pattern, also reported in prior studies~\cite{lin2025vtw, li2025todre}, provides strong evidence that a substantial portion of visual tokens can be pruned without harming performance. 

In contrast, answer tokens in dMLLMs maintain a much stronger attention ratio toward visual tokens across different generation lengths and model backbones. In from-scratch dMLLMs, this ratio remains consistently high throughout dLLM layers (orange line in \Cref{fig:observation} (e-f)), whereas in AR-to-diffusion dMLLMs, it displays a sharp spike at a specific layer (orange line in \Cref{fig:observation} (d)). This strong visual dependence explains the pronounced performance degradation observed in \Cref{tab:lladav} and \Cref{tab:lavida} when pruning is applied, since removing key visual information that answer tokens rely heavily on tends to disrupt the inference process. This phenomenon is consistent with our experimental results for all tasks and model types, except for the long-answer setting in LLaDA-V, which will be discussed in the following subsection. \looseness=-1

\subsection{Restoration Ability Under Information Loss}
\label{subsec:restoration_ability}

\textbf{Takeaway: }\textit{From-scratch dMLLMs demonstrate a strong capacity to recover lost visual information through contextual integration, whereas AR-to-diffusion dMLLMs lack such flexibility, retaining an autoregressive tendency that limits their ability to recover from loss. The restoration ability of dMLLMs depends on both the self-attention strength among answer tokens and the number of denoising steps.}

A key distinction between dMLLMs and MLLMs lies in their decoding dynamics \cite{nie2025llada, ye2025dream}. 
Unlike MLLMs, which generate tokens autoregressively and freeze each token once produced, dMLLMs continuously perform bidirectional interactions and information fusion among all answer tokens throughout the denoising steps.
This distinction is reflected in our observations: in from-scratch dMLLMs, answer tokens exhibit the strongest self-attention (red line in \Cref{fig:observation} (e-g)), whereas in AR-to-diffusion dMLLMs, the self-attention is weaker yet still higher than that in MLLMs (red line in \Cref{fig:observation} (d, h)).
Such enhanced self-focus enables dMLLMs to mitigate the information loss caused by pruning visual tokens through continuous information exchange among answer tokens.

For long-answer tasks, this effect can explain our divergent experimental results in \Cref{tab:decoding-steps}. A strong self-focus among answer tokens allows from-scratch dMLLMs to recover from pruning, while a weaker self-focus results in persistent degradation in AR-to-diffusion dMLLMs.
This pattern can be further confirmed by examining cases from DocVQA \cite{mathew2021docvqa} and InfoVQA \cite{mathew2022infographicvqa}, where we observe a consistent pattern unique to from-scratch dMLLMs: a sharp rise in answer-to-answer self-attention between denoising steps often precedes a surge in token confidence (e.g., the increase in logits at Step 8 in \Cref{fig:observation} (c) coincides with the rise in self-attention at Step 7 in \Cref{fig:observation} (f)). For tasks with longer generations, such as Video Detail Captioning \cite{lmmslab2024videodetailcaption}, the same pattern persists but manifests more smoothly—without the sharp spike (\Cref{fig:observation} (b and e)).

For short-answer tasks, aside from attention strength among answer tokens, the number of denoising steps is crucial for understanding the experimental results in \Cref{tab:decoding-steps}. Since short-answer tasks involve very few total denoising steps, the correct answer token (e.g., a number, “Yes/No,” or a single choice) is typically determined within the first one or two denoising steps. Applying pruning too early leaves insufficient room for subsequent denoising to recover the missing information, leading to severe performance drops. Conversely, applying pruning at later steps yields only marginal efficiency gains. Both cases lead to a suboptimal trade-off between performance and efficiency.

Together, these findings show that effective restoration requires both strong answer-to-answer self-attention and a sufficient number of later denoising steps. This explains why visual redundancy emerges only in from-scratch dMLLMs when handling long-answer tasks.

\subsection{Visual Token Redundancy in dMLLMs}
\label{subsec:redundancy}

\textbf{Takeaway: }\textit{Unlike in MLLMs, where redundancy manifests as unnecessary visual tokens that incur extra computational overhead, redundancy in dMLLMs arises from the model’s ability to recover from missing information through iterative bidirectional refinement.}

In MLLMs, visual redundancy typically refers to the unnecessary computational overhead of dispensable visual tokens, which contribute little to the inference process and can therefore be safely pruned without affecting performance.

In contrast, visual redundancy in dMLLMs reflects a fundamentally different and dynamic property. 
Since pruning inevitably causes information loss and dMLLMs can recover part of it through later bidirectional updates, visual redundancy thus describes how much computation can be reduced by leveraging the model’s ability to compensate for missing information while maintaining performance.

Viewed from this perspective, visual redundancy in dMLLMs is tightly coupled with the model’s restoration ability discussed in \Cref{subsec:restoration_ability}. 
When models exhibit strong self-attention among answer tokens, as observed in from-scratch dMLLMs, they can reintegrate dispersed visual cues to reconstruct missing information, leading to a more resilient response to pruning. 
Conversely, when such self-attention is weak, as in AR-to-diffusion dMLLMs, visual redundancy diminishes because the model cannot sufficiently recover pruned information, and even a small reduction in visual tokens leads to noticeable degradation. 

\subsection{Insights for Visual Pruning in dMLLMs}
\label{subsec:insight}

\textbf{Takeaway: }\textit{For from-scratch dMLLMs, both attention scores and logits can serve as indicators of when and where to prune, with progressive pruning or pruning applied after certain decoding steps being more suitable. For AR-to-diffusion dMLLMs, attention scores can serve as an effective metric, suggesting layer-skipping strategies.}

Given the preceding analysis and our definition of visual redundancy, we attribute the absence of visual token redundancy in short-answer tasks for from-scratch dMLLMs to their strong reliance on visual information and the limited number of denoising steps, which together constrain their ability to restore missing information. For long-answer tasks, we observe a consistent correlation between the attention ratio and logits, as mentioned in \Cref{subsec:restoration_ability}. As the attention ratio shifts, the logits also vary accordingly, suggesting that both signals can jointly guide when and how to apply progressive pruning or pruning after certain decoding steps.

For AR-to-diffusion adapted models, another approach is to prevent information loss. Answer tokens exhibit consistently low attention to visual tokens across both short- and long-answer tasks, except for the sharp spike (the orange line at layer 4 in \Cref{fig:observation} (d)). This pattern suggests a less harmful strategy: allow visual tokens to bypass layers with persistently low attention, while preserving those layers around the spike. In practice, this corresponds to attention-aware skipping that excludes visual tokens from low-dependency regions, leveraging the model’s limited restoration ability without excessive information loss.

\section{Conclusion}
\label{sec:conclusion}

We present the first comprehensive analysis of visual token redundancy in dMLLMs. Our study reveals that visual redundancy in dMLLMs is fundamentally different from that in MLLMs: it reflects the model's capacity to recover pruning-induced information loss through iterative bidirectional refinement, rather than mere token-level dispensability. This property emerges only in from-scratch dMLLMs on long-answer tasks, where strong answer-token self-attention and sufficient denoising steps enable effective restoration. These insights lead to practical pruning guidelines: layer-skipping for AR-to-diffusion models and progressive or late-step pruning for from-scratch architectures.

\section*{Acknowledgment}
\begin{flushleft}
This study is funded by the Ministry of Education, Singapore, under the Tier-2 project scheme with project number MOET2EP20123-0003.
\end{flushleft}
{
    \small
    \bibliographystyle{ieeenat_fullname}
    \bibliography{main}
}

\clearpage
\setcounter{page}{1}
\setcounter{section}{0} 
\setcounter{table}{0} 
\setcounter{figure}{0}
\maketitlesupplementary

\section{Experimental Details}
\label{app:exp_setting}

\subsection{Benchmarks}
\label{app:benchmarks}

We conduct our experiments across a diverse suite of multimodal benchmarks. For image understanding, we evaluate on ten datasets: MME \cite{Fu2023mme}, SQA \cite{lu2022sqa}, GQA \cite{hudson2019gqa}, POPE \cite{li2023pope}, MMB \cite{liu2024mmbench}, TVQA \cite{singh2019textvqa}, CQA \cite{masry2022chartqa}, MMMUP \cite{yue2024mmmupro}, IVQA \cite{mathew2022infographicvqa}, and DVQA \cite{mathew2021docvqa}. For video understanding, we further assess performance on two benchmarks: VMME \cite{fu2024videomme} and Video Detail Caption \cite{lmmslab2024videodetailcaption}.

\paragraph{MME.}
MME is a comprehensive benchmark evaluating multimodal models on 14 perception and cognition subtasks, including OCR, counting, spatial localization, and visual recognition of scenes, landmarks, and artworks. All tasks are formulated as binary judgment questions with curated instruction–answer pairs to ensure fairness. We report the standard perception score on 2,374 image–question pairs.

\paragraph{SQA.}
ScienceQA evaluates multimodal reasoning and zero-shot generalization in scientific domains. It covers natural, language, and social sciences, with questions organized hierarchically across multiple topics and skills. Each question is a multiple choice question, often paired with an illustrative image. We evaluate on the image dataset of 2,017 question–answer pairs.

\paragraph{GQA.}
GQA evaluates structured visual reasoning using images, scene graphs, and automatically generated questions. Each image is paired with a scene graph from the Visual Genome dataset \cite{Krishna2016visualgc}, providing detailed objects, attributes, and relations. We follow standard protocol and report accuracy on the test-dev set with 12,578 image–question pairs.

\paragraph{POPE.}
POPE evaluates object hallucination in vision–language models using binary questions about object presence in images from the MSCOCO dataset \cite{Lin2014microsoftcc}. Performance is measured by the average F1 score over three sampling strategies, covering 8,910 image–question pairs.

\paragraph{MMB.}
MMBench provides a hierarchical evaluation of multimodal understanding across three levels\textemdash{}perception and reasoning (L1), six sub-skills (L2), and 20 tasks (L3)\textemdash{}each formulated as multiple-choice questions. It is available in English and Chinese versions, containing 4,377 and 4,329 image–question pairs, respectively. We evaluate on MMBench-EN subset.

\paragraph{TVQA.}
TextVQA benchmarks VQA models that must read and reason over text in natural images. It comprises 45,336 questions on 28,408 images (from text-rich Open Images categories), with 10 human answers per question. We follow the standard setting and evaluate accuracy on this dataset.

\paragraph{CQA.}
ChartQA benchmarks question answering over chart images that require both visual and logical reasoning. It includes 9,608 human-written questions and 23,111 questions generated from chart summaries, spanning 20,882 real-world charts collected from Statista, Pew Research, Our World in Data, and the OECD. Answers are often open-vocabulary and may involve arithmetic or comparisons. We follow the dataset’s official evaluation protocol.

\paragraph{MMMUP.}
MMMU-Pro is a strengthened version of MMMU that aims to test genuine multimodal understanding and reasoning. It (i) filters out items solvable by text-only models, (ii) augments candidate options, and (iii) introduces the vision-only input setting MMMU-Pro Vision, where questions and options are embedded directly into images so models must truly “see” and read. In our experiments, we report results on the Vision subset following the paper’s protocol.

\paragraph{IVQA.}
InfoVQA evaluates VQA on infographics that require joint reasoning over layout, embedded text, graphical elements, and data visualizations. The dataset contains 5,485 images with 30,035 questions; answers are mainly extractive, with some numerical ones derived via counting, sorting, or simple arithmetic. We follow the official protocol and report accuracy.

\paragraph{DVQA.}
DocVQA focuses on question answering over real document images that require both reading and layout understanding. The dataset contains 12,767 document images of varied types and content, paired with about 50,000 human-annotated question–answer pairs. Each question involves information extraction, reasoning across text blocks, or interpreting document structure. We follow the dataset’s official evaluation protocol.

\paragraph{VMME.}
VideoMME is a large-scale benchmark for evaluating video understanding in LVLMs. It includes 900 videos ($\approx$254 hours) from six domains and 30 subcategories, covering short ($\leq$2 min), medium (4–15 min), and long (30–60 min) durations. Each video has three expert-authored multiple-choice questions, yielding 2,700 video–question pairs. We evaluate on the full dataset.

\paragraph{VDC.}
Video Detail Caption is a video captioning benchmark released by LMMs-Lab, where each video clip is paired with a detailed textual description. The test set contains 499 samples, each including a video name, a question prompt, and an answer paragraph. We follow the official evaluation protocol and assess performance using GPT-4o-mini as the evaluator.

\subsection{Backbone Models}
\label{app:backbones}

\paragraph{LLaDA-V.}
LLaDA-V \cite{you2025lladav} represents the pure diffusion paradigm in multimodal large language modeling. It extends the LLaDA diffusion language backbone with a SigLIP-2 vision encoder and a lightweight MLP projector, enabling multimodal understanding entirely through masked diffusion rather than next-token prediction. As a purely diffusion-trained model, LLaDA-V exemplifies non-autoregressive probabilistic reasoning and demonstrates strong scalability across image, document, and video understanding benchmarks.

\paragraph{LaViDa-Dream.}
LaViDa-Dream \cite{li2025lavida} represents the autoregressive-to-diffusion adaptation paradigm. It builds on Dream-7B, a discrete diffusion language model (DLM) adapted from autoregressive pretraining, and extends it to the multimodal setting through visual instruction tuning. By incorporating techniques such as complementary masking and Prefix-DLM caching, LaViDa-Dream achieves efficient multimodal reasoning while exemplifying the AR-to-diffusion adaptation route in dMLLMs.

\subsection{Token Compression Methods}
\label{app:compression_methods}

\paragraph{ToMe.}
ToMe \cite{bolya2022tome} is a training-free efficiency method that accelerates inference by merging similar tokens instead of pruning them. It computes pairwise similarity between attention keys and merges the most redundant token pairs during encoding through a fast bipartite matching algorithm.

\paragraph{DivPrune.}
DivPrune \cite{alvar2025divprune} formulates visual token pruning as a diversity-driven token selection problem. It defines a min–max diversity objective, encouraging the retained tokens to be maximally dissimilar to each other, and applies a greedy selection strategy to iteratively preserve the most informative and diverse subset of visual tokens.

\paragraph{FastV.}
FastV \cite{chen2024fastv} is a training-free method that accelerates vision–language models by pruning redundant visual tokens in the early decoding stage. It removes the least informative tokens after the second LLM layer based on averaged attention scores.

\paragraph{VTW.}
VTW \cite{lin2025vtw} is a training-free acceleration method that withdraws all visual tokens after a specific transformer layer to reduce inference cost in vision–language models. The withdrawal layer is chosen via a KL divergence criterion, enabling VTW to cut FLOPs and memory usage by over 40\% without significant performance degradation.

\paragraph{SparseVLM.}
SparseVLM \cite{zhang2024sparsevlm} introduces adaptive cross-modal sparsity to reduce redundancy in both visual and textual tokens.
It ranks token importance via cross-modal attention, dynamically applies different sparsity ratios to vision and language streams, and employs a token recycling mechanism that reuses informative pruned tokens to preserve contextual completeness.

\paragraph{TRIM.}
TRIM \cite{ruiz2024trim} is a training-free token reduction method that measures text–image similarity in the CLIP representation space to rank visual tokens. It selects important tokens via an IQR-based threshold and appends an aggregated representation of unselected tokens.

\section{Generation Hyperparameters}
\label{app:hyperparameters}

\begin{table}[t]
\centering
\fontsize{10pt}{11.5pt}\selectfont
\setlength{\tabcolsep}{3.5pt}
\renewcommand{\arraystretch}{1.08}
\adjustbox{max width=\columnwidth}{
\begin{tabular}{lcccc}
\toprule
\textbf{Benchmark(s)} & \textbf{gen\_length} & \textbf{block\_length} & \textbf{gen\_steps} & \textbf{think\_mode} \\
\midrule
\multicolumn{5}{c}{\textbf{LLaDA-V} \cite{you2025lladav}} \\
\midrule
\makecell[l]{MME, SQA, GQA,\\ POPE, MMB, TVQA,\\ MMMUP, VMME} & 2 & 2 & 2 & no\_think \\
CQA & 16 & 16 & 8 & no\_think \\
DVQA, IVQA & 32 & 32 & 16 & no\_think \\
VDC & 128 & 128 & 64 & think \\
\midrule
\multicolumn{5}{c}{\textbf{LaViDa-Dream} \cite{li2025lavida}} \\
\midrule
\makecell[l]{MME, SQA, GQA,\\ POPE, MMB, TVQA,\\ MMMUP} & 4 & 4 & 2 & no\_think \\
CQA & 16 & 16 & 8 & no\_think \\
DVQA, IVQA & 32 & 32 & 16 & no\_think \\
\bottomrule
\end{tabular}}
\caption{\textbf{Generation hyperparameters used for different benchmarks.}
The settings largely follow the default configurations of the respective backbone models, with minor adjustments to ensure stable decoding across short- and long-answer tasks.}
\label{tab:gen_kwargs}
\vspace{-2mm}
\end{table}

We summarize in \Cref{tab:gen_kwargs} the generation hyperparameters used in our experiments for both the LLaDA-V and LaViDa-Dream backbone models. The generation settings generally follow the default configurations provided in the original model implementations, with minor adjustments to ensure stable decoding across short- and long-answer tasks. Specifically, for LaViDa-Dream, the parameters \texttt{gen\_length}, \texttt{block\_length}, and \texttt{gen\_steps} are set to 4, 4, and 2, respectively, for the group of benchmarks including MME, ScienceQA, and GQA to ensure decoding stability.

\end{document}